%% file: main.tex
\documentclass[letterpaper, 10 pt, journal, twoside]{IEEEtran}

\usepackage{adjustbox}
\usepackage{amsmath}
\usepackage{amssymb}
\usepackage{booktabs}
\usepackage{comment}
\usepackage{float}
\usepackage{graphicx}
\usepackage{mathtools}
\usepackage{multirow}
\usepackage{multicol}
\usepackage{subfig}
\usepackage{xspace}
\usepackage{xfrac}
\usepackage{xfrac}
\usepackage{pifont}
\usepackage[table]{xcolor}

\definecolor{darkgreen}{RGB}{0, 150, 0}
\definecolor{darkred}{RGB}{200, 0, 0}
\newcommand{\ch}{{\color{darkgreen} \ding{51}}}
\newcommand{\xm}{{\color{darkred} \ding{55}}}

\usepackage[colorlinks,pagebackref=false,citecolor=green,bookmarks=false,hypertexnames=true]{hyperref} 

\DeclarePairedDelimiter{\abs}{\lvert}{\rvert}

\newcommand{\lone}{$L_1$\xspace}
\newcommand{\ltwo}{$L_2$\xspace}

\makeatletter
\DeclareRobustCommand\onedot{\futurelet\@let@token\@onedot}
\def\@onedot{\ifx\@let@token.\else.\null\fi\xspace}
\def\eg{\emph{e.g}\onedot} 
\def\ie{\emph{i.e}\onedot} 
\def\etal{\emph{et al}\onedot}
\makeatother
\begin{document}
\title{SVDistNet: Self-Supervised Near-Field Distance Estimation on Surround View Fisheye Cameras}

\author{Varun~Ravi~Kumar,
        Marvin~Klingner,
        Senthil~Yogamani,
        Markus~Bach, \\
        Stefan~Milz,
        Tim~Fingscheidt,
        and Patrick~Mäder
\thanks{V. Ravi Kumar, M. Bach are with Valeo DAR Kronach, Kronach, Germany.}
\thanks{V. Ravi Kumar, S. Milz and P. Mäder are with TU Ilmenau, Germany.}
\thanks{M. Klingner and T. Fingscheidt are with TU Braunschweig, Germany.}
\thanks{S. Yogamani is with Valeo Vision Systems, Tuam, Ireland.}
\thanks{Manuscript received September 13, 2020}}

\markboth{IEEE Transactions on Intelligent Transportation Systems}
{}
\maketitle
\begin{abstract}
A $360^\circ$ perception of scene geometry is essential for automated driving, notably for parking and urban driving scenarios. Typically, it is achieved using surround-view fisheye cameras, focusing on the near-field area around the vehicle. The majority of current depth estimation approaches focus on employing just a single camera, which cannot be straightforwardly generalized to multiple cameras. The depth estimation model must be tested on a variety of cameras equipped to millions of cars with varying camera geometries. Even within a single car, intrinsics vary due to manufacturing tolerances. Deep learning models are sensitive to these changes, and it is practically infeasible to train and test on each camera variant. As a result, we present novel camera-geometry adaptive multi-scale convolutions which utilize the camera parameters as a conditional input, enabling the model to generalize to previously unseen fisheye cameras. 
Additionally, we improve the distance estimation by pairwise and patchwise vector-based self-attention encoder networks. We evaluate our approach on the Fisheye WoodScape surround-view dataset, significantly improving over previous approaches. We also show a generalization of our approach across different camera viewing angles and perform extensive experiments to support our contributions. To enable comparison with other approaches, we evaluate the front camera data on the KITTI dataset (pinhole camera images) and achieve state-of-the-art performance among self-supervised monocular methods. An overview video with qualitative results is provided at {\url{https://youtu.be/bmX0UcU9wtA}}. Baseline code and dataset will be made public\footnote{\url{{https://github.com/valeoai/WoodScape}}}.
\end{abstract}
\IEEEpeerreviewmaketitle
\input{include/introduction}
\input{include/related_work}
\input{include/method}
\input{include/network_details}
\input{include/implementation_details}
\input{include/experiments}
\section{Conclusion}
\label{sec:conclusion}

In this paper, we solved the problem of multi-camera distance estimation for surround-view fisheye cameras. We introduced a novel camera model adaptation mechanism wherein camera parameters are transformed into a tensor and used within the CNN model. The specific camera model parameters are used during training and inference. Using this technique, we demonstrate training of a single distance estimation model for twelve different cameras with different extrinsic and intrinsic parameters and achieve the improved results as training a specialized model for each camera variant. We achieve state-of-the-art results on the fisheye WoodScape dataset and the KITTI dataset. We intend to learn and refine camera parameters' within the training framework in future work.\par
\bibliographystyle{IEEEtran}
\bibliography{IEEEabrv,bib/egbib}
\input{include/biographies}
\end{document}

%% file: include/introduction.tex
\section{Introduction}
\begin{figure}[t]
  \captionsetup{singlelinecheck=false, font=small,  belowskip=-15pt}
  \centering
    \includegraphics[width=0.95\linewidth]{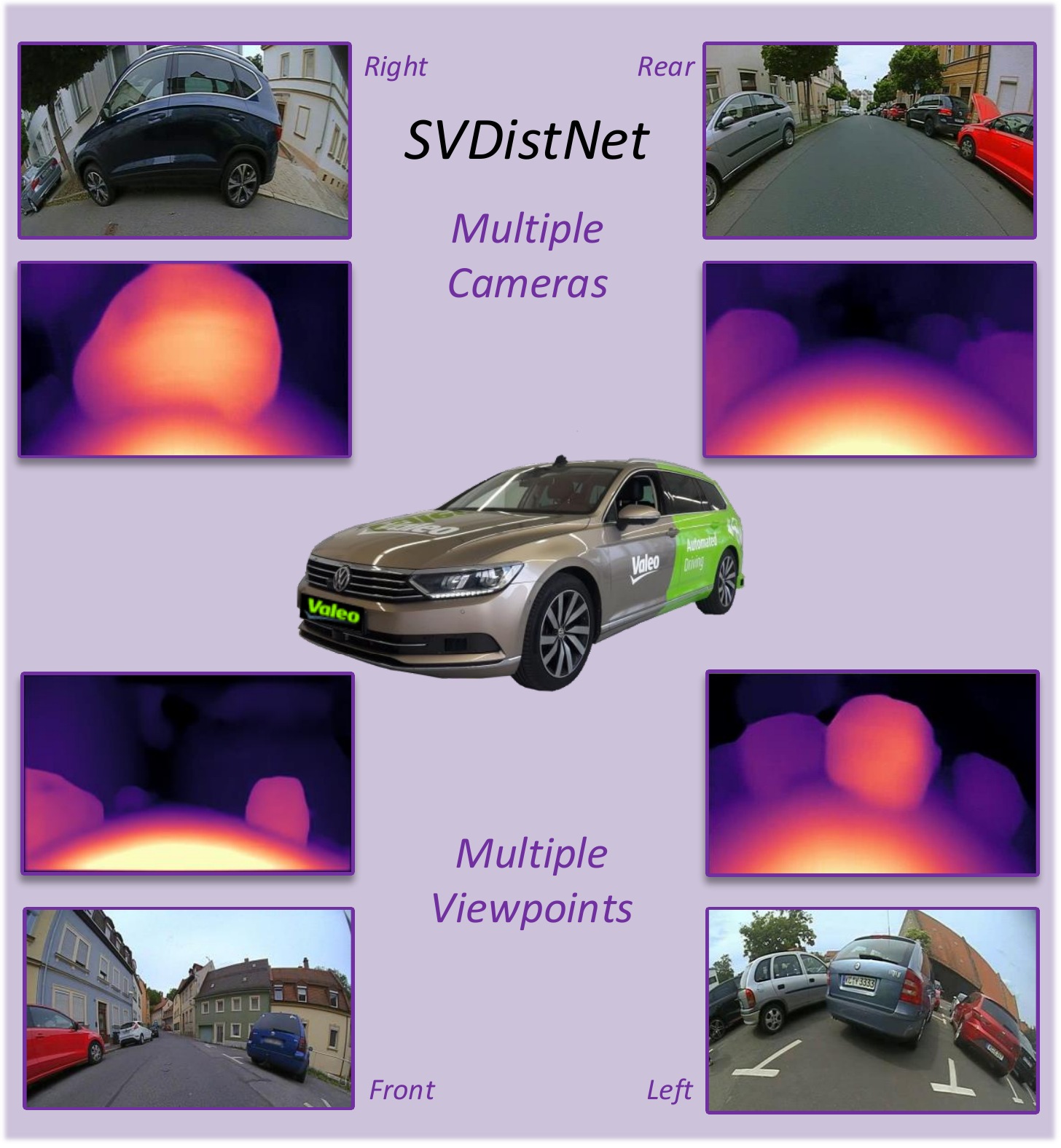} \caption{\textbf{Our surround-view distance estimation framework} is facilitated by employing a single network on images from multiple cameras. A surround-view coverage of geometric information is obtained for an autonomous vehicle by utilizing and post-processing the distance maps from all cameras.}
    \label{fig:general_concept}
\end{figure}
\textbf{\textit{Surround View Cameras}} Typically, automotive perception systems use multiple cameras, with current systems having at least four cameras. The number is likely to increase to more than ten cameras for future generation systems. Such surround-view cameras are focused on near field sensing, which is typically used for low-speed applications such as parking or traffic jam assistance functions~\cite{heimberger2017computer}. Fig.~\ref{fig:general_concept} shows sample distance estimation images of four cameras mounted on a car covering the entire 360$^{\circ}$ field of view surrounding the car. Near-field distance estimation is a challenging problem because of distortion and partial visibility of close-by objects~\cite{kumar2018monocular}. Also, centimeter-level accuracy is required to enable precise low-speed maneuvers such as parking. Up to now, for the generation of high-quality distance estimates, one network per camera has to be trained, inducing unfeasible computational complexity with an increasing number of cameras.\par
\textbf{\textit{Motivation of SVDistNet}} Our work's primary motivation is to propose a solution for surround-view fisheye cameras targeting large-scale industrial deployment. In other words, we aim to develop a model that is applicable to a vast number of automated driving vehicles possessing their own set of cameras.  However, the underlying camera intrinsics are not even entirely identical for a given vehicle family. There are differences due to manufacturing processes that lead to a separate calibration of each camera instance. Calibration can vary even after deployment due to high environmental temperatures or aging. As a result, we require a calibration adaptation mechanism in the model. This contrasts with public datasets, which employ a single camera instance for both training and testing. There are 12 different cameras with slight intrinsic variations in the Woodscape dataset~\cite{yogamani2019woodscape} to assess this effect. There are 4 camera instances around the vehicle with varying intrinsics, even for a single instance of a surround-view system. Instead of having four separate models for each camera, we, therefore, propose a single model for all cameras, which has advantages such as (i) a lower memory footprint and a better data transmission rate leading to improved efficiency, (ii) a better regularization during training due to the wider variety of different cameras in a more extensive training dataset, and (iii) an easier to handle maintenance of just a single neural network model instead of multiple ones.
\par
Recently proposed methods for self-supervised depth estimation~\cite{zhou2017unsupervised, Godard2019, kumar2021fisheyedistancenet++} present a powerful tool for the perception of scene geometry from camera images as they are trainable on arbitrary image sequences. While initial methods mainly focused on pinhole cameras~\cite{Casser2019, Godard2019, Klingner2020a}, for practical deployment, the more general distance estimation~\cite{kumar2020fisheyedistancenet, kumar2020unrectdepthnet} on other camera geometries such as fisheye cameras are of significant interest due to their large field of view. There are many benefits in using raw images instead of undistortion, refer to \cite{kumar2020unrectdepthnet} for details. Moreover, current methods essentially focus on single front-view camera systems, while the question of \emph{how to extend such models to multiple cameras is essentially left open}. Therefore, in this work, we extend our previous synergized multi-task learning method for self-supervised distance estimation and semantic segmentation~\cite{kumar2020syndistnet} to a general application on surround-view cameras and propose additional improvements.\par
To sum up, our contributions are as follows:
\begin{itemize}
    \item We present a novel camera geometry adaptive multi-scale convolution to incorporate the camera parameters into our self-supervised distance estimation framework. We feed this camera geometry tensor representation to the model as a generic way to adapt to new camera intrinsics.
    \item We create a training framework for self-supervised distance estimation, which jointly trains and infers images from multiple fisheye cameras and viewpoints. 
    \item We demonstrate a single model for 12 fisheye cameras, which achieves the equivalent result as an individual specialized model that overfits a particular camera model.
    \item We present an improved version of our network architecture for multi-task learning of self-supervised distance estimation and semantic segmentation. We significantly improve upon our previous works FisheyeDistanceNet~\cite{kumar2020fisheyedistancenet}, UnrectDepthNet~\cite{kumar2020unrectdepthnet}, and SynDistNet~\cite{kumar2020syndistnet}.
    \item We achieve state-of-the-art results on the WoodScape and KITTI datasets among monocular self-supervised depth estimation methods.
\end{itemize}\par

%% file: include/related_work.tex
\section{Related Work}
\label{sec:related_work}
This section provides a survey regarding self-supervised depth estimation and, in particular, its applications to different camera geometries. Also, we discuss the application of self-supervised depth estimation in multi-task learning approaches and approaches using semantic guidance for this task.\par
\textbf{\textit{Self-Supervised Depth Estimation}}
Training a neural network to predict a depth map by optimizing this depth map as a parameter of geometric projections between stereo images or sequential images has been defined as the task of self-supervised depth estimation~\cite{zhou2017unsupervised}. For performance gains, the following approaches extended the fundamental concept by improved loss functions~\cite{Godard2017, Godard2019}, or network architectures, optimized for the task of depth estimation~\cite{Guizilini2020a}. Concurrently, it was proposed to use proxy labels from traditional stereo algorithms~\cite{Tosi2019}. To better model the structural information from the temporal context, recurrent neural networks were employed~\cite{Wang2019, Zhang2019c}, and Casser~\etal~\cite{Casser2019, Casser2019a} introduced a test-time refinement method to improve the network performance in an online manner. While all of these approaches achieve significant results on pinhole camera images, they are seldom transferred to more complex camera models, \eg, fisheye camera images. In contrast, we exhibit our self-supervised distance estimation method's applicability on both pinhole and fisheye camera images.\par
\textbf{\textit{Multi-Task Learning}} Exemplary tasks, which benefit from multi-task learning, \ie, share some of the network parts, in terms of performance and efficiency are, \eg, depth estimation~\cite{Kendall2018, Wang2020a}, semantic segmentation~\cite{Klingner2020, Klingner2021, Wang2020a} and domain adaptation~\cite{Zhao2019}. As an empirical weighting of the losses involves extensive hyperparameter tuning, we follow the uncertainty-based weighting of Kendall~\etal~\cite{Kendall2018} to weigh the different tasks. To handle dynamic objects during training that violate the static world assumption in the projections of the self-supervised depth estimation, it is common to additionally estimate optical flow, which can also be trained in a self-supervised fashion~\cite{Liu2019}. Through this simultaneous estimation of both outputs, it is thereby plausible to enforce cross-task consistency~\cite{Liu2019a}, to modify the projected image and correct the influence of dynamic objects~\cite{Chen2019b}, or to enforce prior known geometric constraints. Although optical flow is the commonly used technique to handle dynamic objects, we consider an alternative method, introduced by~\cite{Klingner2020a, kumar2020syndistnet} because, firstly, semantic segmentation is a task that is readily available in most autonomous driving systems and, secondly, optical flow is computationally more complex and hard to validate due to the extensive effort needed to obtain ground truth.\par
\textbf{\textit{Semantically-Guided Depth Estimation}}
Complementing self-supervised depth estimation by the additional prediction of semantic or instance segmentation in cross-task guidance approaches has been shown to increase the performance of both prediction modalities~\cite{Casser2019, Casser2019a, Klingner2020a, nekrasov2019real, miclea2019real, Guizilini2020}. Either the segmentation masks are given as an additional input to the network~\cite{Guizilini2020}, or they are used to predict relative poses for the single dynamic objects to counterpoise their movement between consecutive frames~\cite{Casser2019, Casser2019a}. While primary approaches used a pre-trained segmentation network, it was also shown that both tasks could be trained simultaneously~\cite{Chen2019a, Zhu2020}. Thereby, it is possible to project the segmentation outputs between consecutive frames and enforce temporal semantic consistency~\cite{Chen2019a, Yang2018b}, or enforce edge consistency across the different output modalities \cite{Chen2019a, Zhu2020}. In this work, we use the segmentation output to identify dynamic objects and handle them accordingly inside the photometric loss as in~\cite{Klingner2020a, kumar2020syndistnet}. Additionally, we propose a multi-task network, where the encoder is based on pairwise and patchwise self-attention networks together with pixel-adaptive convolutions inside the decoder (as in~\cite{Guizilini2020}), which can be trained in a single-stage manner, eliminating the need for a pre-trained segmentation network compared to~\cite{Guizilini2020}.\par
\textbf{\textit{Depth Estimation for Different Camera Geometries}}
Employing self-supervised depth estimation techniques to different camera geometries than a single pinhole camera is challenging for two reasons. Firstly, the intrinsic camera parameters are assumed to be constant for all images. Secondly, the viewing angle is assumed to be constant for all camera images, as usually only a front-facing camera is considered (\eg, \cite{zhou2017unsupervised, Godard2019}). Gordon~\etal~\cite{Gordon2019} predicts not only the relative pose but also the intrinsic camera parameters, enabling the generalization across images from different pinhole cameras. Meanwhile, Facil~\etal~\cite{Facil2019} introduces a method to give the intrinsic camera parameters an additional input to the convolution operation, showing the same effect. In this work, we transfer the approach~\cite{Facil2019} to fisheye camera images, thereby enabling generalization of our distance estimation method across different fisheye cameras and viewing angles.\par
The transfer of self-supervised depth estimation to the more general self-supervised distance estimation on camera geometries, such as, \eg, fisheye cameras, was introduced by Ravi Kumar \etal~\cite{kumar2020fisheyedistancenet, kumar2020unrectdepthnet}, which focus on front-view cameras. Surround-view depth coverage has also been shown by~\cite{Wang2020b, Jin2020}, which applies self-supervised depth estimation techniques to $360^\circ$ images. In this work, we do not have to impose a $360^\circ$ image in advance to the model; we enable a surround-view coverage of geometric information for an autonomous vehicle by applying the same neural network to many arbitrary cameras with different viewing angles.\par

%% file: include/method.tex
\section{Multi-Task Learning Framework}
\label{sec:method}
\begin{figure*}[t]
  \captionsetup{singlelinecheck=false, font=small, belowskip=-12pt}
  \centering
    \includegraphics[width=\textwidth]{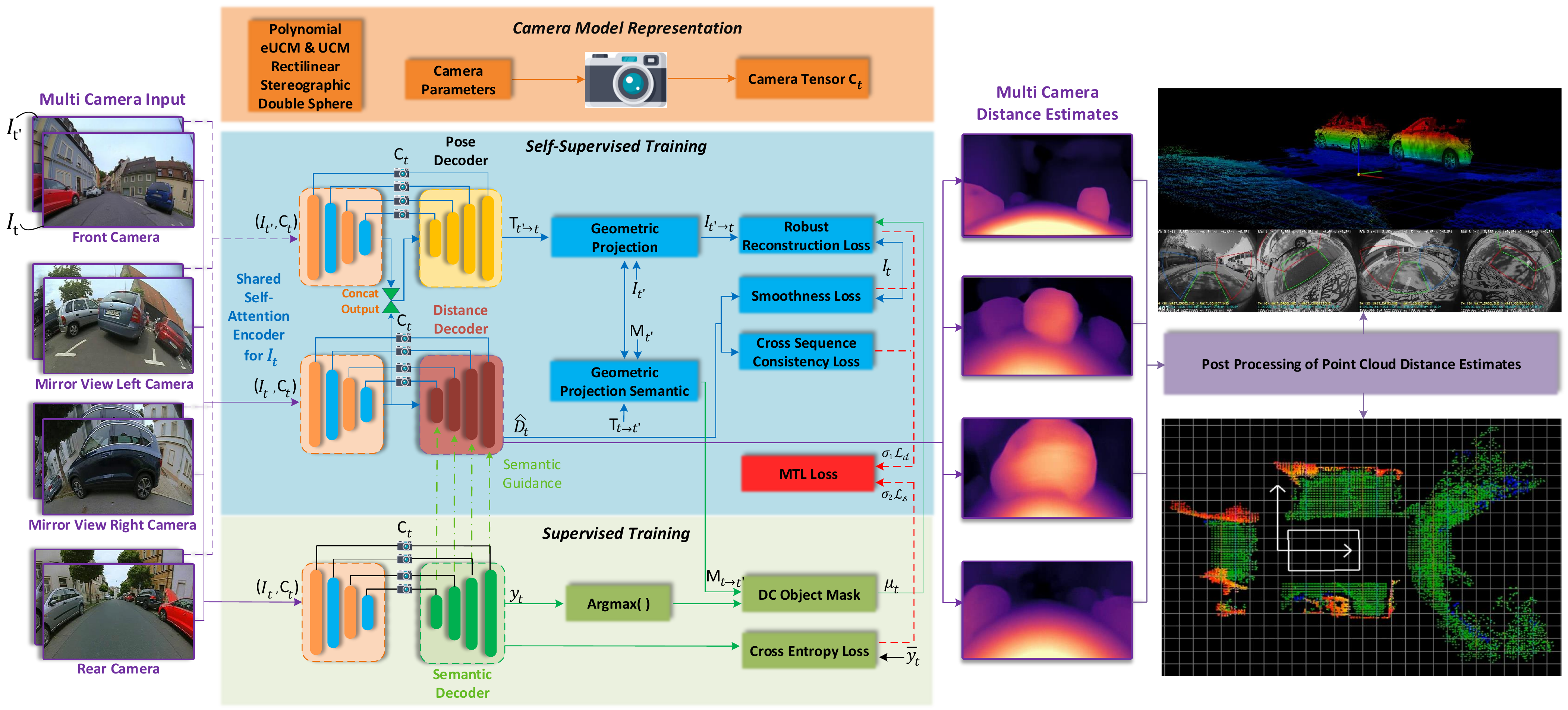}
    \caption{\textbf{High-level overview of our surround-view self-supervised distance estimation framework}, which employs semantic guidance as well as camera-geometry adaptive convolutions (orange blocks). Our framework comprises training units for self-supervised distance estimation (blue blocks) and semantic segmentation (green blocks). The camera tensor $C_t$ (orange block) assists our SVDistNet in producing distance maps across multiple camera-viewpoints and making the network camera independent. As explained in Section~\ref{sec:camera-geometry-tensor} $C_t$ can also be applied to standard camera models. The multi-task loss from~\ref{eq:mtl_loss} weights and optimizes both modalities at the same time. By post-processing the predicted distance maps in 3D space, we can obtain surround-view geometric information using our proposed framework.}
    \label{fig:mtl_pipeline}
\end{figure*}
This section describes our multi-task learning framework for collaborative distance and semantic segmentation prediction, which we apply to surround-view camera systems. 
The learning framework is based on our previous work
FisheyeDistanceNet~\cite{kumar2020fisheyedistancenet}, UnrectDepthNet~\cite{kumar2020unrectdepthnet}, and SynDistNet~\cite{kumar2020syndistnet}  with extensions to support multiple cameras. To enable comparison, we use the same protocols and losses.
\subsection{\textbf{Self-Supervised Distance and Pose Estimation}}
We develop our single-image distance and pose estimation framework for multiple cameras building on our previous work FisheyeDistanceNet~\cite{kumar2020fisheyedistancenet}, where we outlined the basic mechanism of our polynomial projection model for fisheye cameras. Accordingly, the total loss for our distance and pose estimation networks (cf.~Fig.~\ref{fig:mtl_pipeline}) consists of a image reconstruction contribution $\mathcal{L}_r$, minimizing the difference between the reconstructed images $\hat{I}_{t'\to t}$ and the target frames $I_t$ and a regularizing contribution $\mathcal{L}_s$, enforcing edge-aware smoothness as outlined in~\cite{Godard2017}. Additionally, we make use of the cross-sequence distance consistency loss $\mathcal{L}_{dc}$ as well as the scale recovery technique from~\cite{kumar2020fisheyedistancenet}. The total loss is averaged over image batches, scales, and pixel locations and is given as  
\begin{align}
    \mathcal{L}_{tot} = \mathcal{L}_r(I_t,\hat{I}_{t'\to t}) + \beta~\mathcal{L}_s(\hat{D}_t) + \gamma~\mathcal{L}_{dc}(\hat{D}_t,\hat{D}_{t'}).
    \label{eq:overall-loss}
\end{align}
with $\beta$ and $\gamma$ being constants, weighing the influence of the regularization term $\mathcal{L}_{s}$ and the cross-sequence distance consistency term $\mathcal{L}_{dc}$, respectively.\par
\textbf{\textit{Image Reconstruction Loss}} 
\label{sec:reconstruction loss}
The majority of cutting-edge self-supervised depth estimation approaches use heuristic loss functions. The optimal selection of a loss function, on the other hand, is not well defined theoretically. We emphasized the importance of investigating a better reconstruction loss in our previous paper, SynDistNet~\cite{kumar2020syndistnet}. We incorporated Barron's~\cite{barron2019general} more generic robust loss function, which was used to replace the \lone term employed in~\cite{kumar2020fisheyedistancenet, kumar2020unrectdepthnet, Godard2017, Godard2019}. In the context of distance/depth estimation, we presented the common notion of a per-pixel regression $\rho$, indicated by
\begin{align}
    \rho \left( \xi \right) = \rho \left(I_t - \hat I_{t' \to t} \right)
\end{align}
This robust loss function generalizes several common losses, including the  \lone, \ltwo, Geman-McClure, Welsch/Leclerc, Cauchy/Lorentzian, and Charbonnier loss functions. Robustness is introduced as a continuous parameter that can be optimized within the loss function to increase regression task performance. The robust loss function $\rho_{\mathrm{rob}}$ is as follows:
\begin{align} 
  \label{equ:robust_loss}
    \rho_{\mathrm{rob}}\left(\xi\right) = \frac{\abs{\alpha - 2}}{\alpha} \left( \left( {\frac{\left( \sfrac{\xi}{c} \right)^2}{\abs{\alpha - 2}}} + 1 \right)^{\sfrac{\alpha}{2}} - 1 \right)
\end{align}
As described in~\cite{barron2019general}, one can use a data-driven optimization to automatically adapt the free parameters $\alpha$ and $c$ to the investigated problem. The image reconstruction loss between the target frame $I_t$ and the reconstructed target frame $\hat I_{t' \to t}$ is computed by a mixture of the Structural Similarity loss term and the robust pixel-wise loss term as
\begin{align}
  \label{eq:loss-photo}
    \rho_{\mathrm{rob}}\left(\xi\right) &= \rho 
    \left\| (I_t - \hat I_{t' \to t}) \odot \mathcal{M}_{t \to t^\prime} \right\| \\
    \tilde{\mathcal{L}}_{r}(I_t,\hat I_{t' \to t}) 
    &= \tau~\frac{1 - \text{SSIM}(I_t,\hat I_{t' \to t}, \mathcal{M}_{t \to t^\prime})}{2} \nonumber \\
    &\quad+ (1-\tau) \rho_{\mathrm{rob}}\left(\xi\right)
\end{align}
where $\tau=0.85$ is the weighting factor between both loss terms and $\mathcal{M}_{t \to t^\prime}$ is the binary ego-mask employed from~\cite{kumar2020fisheyedistancenet}. Finally, we apply the per-pixel minimum reconstruction loss $\mathcal{L}_{r}$~\cite{Godard2019} by a computation over all source images.
\begin{align}
    \mathcal{L}_{r} &= \min_{t^\prime \in \{t+1,t-1\}} \tilde{\mathcal{L}}_r(I_t, \hat I_{t' \to t})
\end{align}\par
\subsection{\textit{\textbf{Semantic Guidance for Image Reconstruction Loss}}}
To improve distance estimation performance through handling dynamic objects accordingly, in this part, we describe how we train our semantic segmentation baseline and afterward describe the semantic masking technique incorporated from our previous work~\cite{kumar2020syndistnet} to exclude their influence on the reconstruction loss.\par
\textbf{\textit{Semantic Segmentation Baseline:}}
Semantic segmentation aims at assigning a class label $s$ from a subset of classes $s\in\mathcal{S} =  \left\lbrace1,2,...,S\right\rbrace$. The desired output is then a pixel-wise semantic segmentation mask $M_{t}$ with the same spatial dimensions as the segmentation network's input $I_t$. During training, the network predicts posterior probabilities $Y_t$, representing the likelihood that a pixel belongs to a class $s\in \mathcal{S}$. A cross-entropy loss is then used to optimize this posterior probability in a supervised manner.
\begin{equation}
\mathcal{L}_{ce} = -\sum_{s \in\mathcal{S}} \overline{Y}_{t,s} \cdot \log\left(Y_{t,s}\right)
\label{eq:crossentropy_loss}
\end{equation}
which is averaged per pixel and where $\overline{Y}_{t}$ represents one-hot encoded ground truth labels. The final segmentation mask $M_t$ is obtained by performing a pixel-wise  $\operatorname{argmax}$ operation on the posterior probabilities $Y_{t,s}$.\par
\textbf{\textit{Dealing With Dynamic Objects:}}
\label{sec:dynamic-object-mask}
Because dynamic objects defy the static world assumption, knowledge of their depth/distance is critical in autonomous driving. Therefore, we use the information provided by the semantic segmentation to learn the distance from \textit{non-moving} dynamic objects, while we exclude potentially \textit{moving} dynamic objects. To this end, we employ a pixel-wise mask $\mu_t$ containing a $1$, if a pixel does neither show a dynamic object from the current frame $I_t$ nor a wrongfully projected dynamic object from the reconstructed frames $\hat{I}_{t'\to t}$. Otherwise, we set the value of the mask to $0$. Accordingly, we have to predict a segmentation mask $M_t$ from the target frame $I_t$, as well as segmentation masks $M_{t'}$ for the source frames $I_{t'}$. While we can easily detect the dynamic objects in the source frame's segmentation mask $M_t$, the wrongfully projected dynamic objects can be obtained by warping the segmentation masks and employing a nearest-neighbor sampling to the target frame. This yields projected segmentation masks $M_{t' \to t}$. Finally, we can define the set of dynamic object classes $\mathcal{S}_{\mathrm{DC}} \subset \mathcal{S}$ and reduce the semantic segmentation mask to a binary mask, with the properties as outlined above by computing its elements at location $ij$ as:
\begin{equation}
\!\!\mu_{t, ij} = 
\!\left\{
\begin{array}{l}
1 ,\; M_{t, ij} \notin \mathcal{S}_{\mathrm{DC}}\; \land \; M_{t'\rightarrow t,ij} \notin \mathcal{S}_{\mathrm{DC}} \\
0,\; \mathrm{else} \\
\end{array}
\right.
\label{eq:semantic_mask}
\end{equation}
We mask dynamic objects by multiplying the mask with the reconstruction loss from Eq.~\ref{eq:loss-photo} in a pixel-wise fashion. However, there are images in which these dynamic objects are not moving, \eg, parking cars are perfect to learn the distance of cars, while cars on a highway severely violate the static world assumption. Accordingly, we want to learn the distance of \textit{moving} dynamic objects by learning from images, in which they are detected as \textit{non-moving}. As a result, we detect images with non-moving dynamic objects using a semantic consistency measure introduced in our previous works~\cite{Klingner2020a, kumar2020syndistnet} and only mask the dynamic objects in a fraction $epsilon$ of images where dynamic objects are detected as mostly moving.
\subsection{\textbf{Joint Optimization}}

Based on Kendall's \etal~\cite{Kendall2018} task weighting strategy, the distance estimation and semantic segmentation loss terms from Eq.~\ref{eq:overall-loss} and Eq.~\ref{eq:crossentropy_loss}, respectively, are weighted. The total optimization objective is derived from~\cite{kumar2020syndistnet}.
\begin{align}
    \frac{1}{2 \sigma_1^2} \mathcal{L}_{tot} + \frac{1}{2 \sigma_2^2} \mathcal{L}_{ce} + \log (1 + \sigma_1) + \log (1+ \sigma_2)
    \label{eq:mtl_loss}
\end{align}
where $\sigma_1$ and $\sigma_2$ are learnable parameters introduced to optimize the homoscedastic uncertainty. The goal is to optimize a more substantial uncertainty, resulting in a smaller contribution of the loss from the distance estimation or semantic segmentation task to the total loss. It is important to note that the respective task's weight is reduced by increasing the noise parameter $\sigma$. We observe that the noise parameter $\sigma_1$ of the distance estimation task is lower than $\sigma_2$ of semantic segmentation, and over time, the convergence occurs correspondingly.\par


%% file: include/network_details.tex
\section{Network Architecture}
\label{sec:network_details}

This section outlines our novel architecture for semantically-guided self-supervised distance estimation utilizing the camera geometry tensor to handle varying camera intrinsics and viewpoints. We also incorporate pairwise and patchwise self-attention encoders from~\cite{zhao2020exploring} which improves on our baseline SynDistNet~\cite{kumar2020syndistnet}, which employs a scalar self-attention encoder. Thereby, an adaptation of the weights across spatial and channel dimensions is implemented. We retain our semantic guidance decoder from~\cite{kumar2020syndistnet} to facilitate the distance decoder to yield accurate distance predictions. The complete training of both tasks is performed in a one-stage manner.\par
\begin{figure}[t]
  \captionsetup{singlelinecheck=false, font=small, belowskip=-14pt}
  \centering
    \includegraphics[width=0.95\columnwidth]{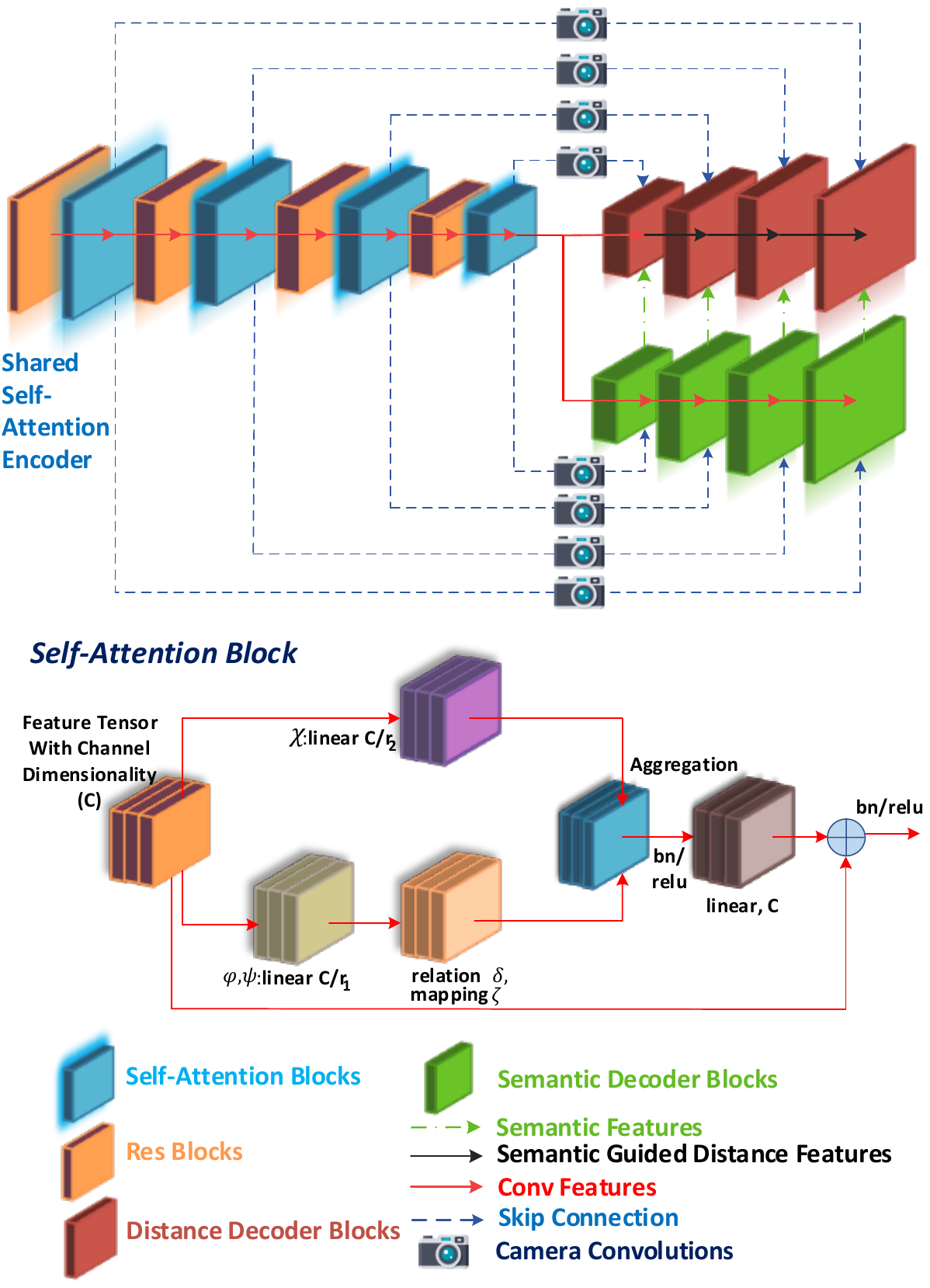}
    \caption{\textbf{Overview of our proposed network architecture} for semantically guided self-supervised distance estimation. It consists of a shared vector-based self-attention encoder and task-specific decoders. Our encoder is a self-attention network with pairwise and patchwise variants, while the decoder uses pixel-adaptive convolutions, which are both complemented by our novel Camera Geometry convolutions.}
    \label{fig:model_arch}
\end{figure}
\subsection{\textbf{Camera Geometry Tensor $C_t$}}
\label{sec:camera-geometry-tensor}

Automated driving systems have a wide variety of cameras, typically around 10, placed in different car locations and with different fields of view. Instead of developing an individual model for each camera, it is highly desirable to develop a single model for all cameras, as discussed in the introduction. This is an unsolved problem, and we aim to solve this by incorporating camera geometry into the distance estimation. We intend to convert all the camera geometry characteristics into a tensor called camera geometry tensor $C_t$ and provide this as input to the CNN model at both training and inference. From the view of distance estimation, camera intrinsics is the primary model adaptation needed. However, the camera tensor notion is generic, and we plan to extend it to include camera extrinsics and camera motion (visual odometry) for improving other tasks. CAM-Convs~\cite{Facil2019}, which utilizes camera-aware convolutions for pinhole cameras, is the closest work. We extend and generalize this work to arbitrary camera geometries, including fisheye cameras.\par

The camera geometry adaptive mechanism is fundamental in the training process of the SVDistNet as the four different cameras mounted on the car have different intrinsic parameters and viewpoints. The trained distance and pose estimation networks need to generalize when deployed on a different car with a change in multiple viewpoints and intrinsics. To accomplish this, as shown in Fig.~\ref{fig:model_arch}, we include the camera geometry tensor $C_t$ in the mapping from RGB features to 3D information for distance estimation and semantic segmentation. In addition, as shown in Fig.~\ref{fig:mtl_pipeline}, we add $C_t$ to the pose encoder.\par
The tensor $C_t$ can be obtained using a three step procedure: 
First, we pre-compute pixel-wise coordinates and the angle of the incidence maps to facilitate a more efficient training. To incorporate the knowledge encoded in these camera calibration parameters, the normalized coordinates per pixel are used as additional channels and concatenated, yielding the tensor $C_t$. Finally, these features are passed along with the encoder features to our SAN \emph{pairwise} and \emph{patchwise} models. Accordingly, we have six additional channels, supplementing the existing decoder channel inputs. In theory, the proposed method can be applied to any given fisheye projection model. We briefly present the standard projection models that our camera geometry tensor supports. The mapping of 3D points to pixels in fisheye lenses requires a radial component $\varrho\left(\vartheta\right)$. The polynomial model is the most commonly used, and more recent projection models include UCM (Unified Camera Model)~\cite{barreto2006unified} and eUCM (Enhanced UCM)~\cite{Khomutenko2016eucm}. Rectilinear (representation of a pinhole model) and Stereographic (mapping of a sphere to a plane) are not appropriate for fisheye lenses but are provided for comparison. The recently proposed Double Sphere model~\cite{usenko2018double} has a closed-form inverse with low computational complexity. The following are summaries of the radial distortion models:
\begin{enumerate}
\item Polynomial: $\varrho(\vartheta) = a_1 \vartheta + a_2 \vartheta^2 + a_3 \vartheta^3 + a_4 \vartheta^4$
\item UCM: $\varrho(\vartheta) = f\cdot\sin\vartheta / (\cos\vartheta + \xi)$
\item eUCM: $\varrho(\vartheta) = f\cdot\frac{\sin\vartheta}{\cos\vartheta + \alpha\left(\sqrt{\beta\cdot \sin^2\vartheta + \cos^2\vartheta} - \cos\vartheta\right)}$ 
\item Rectilinear: $\varrho(\vartheta) = f \cdot \tan\vartheta$
\item Stereographic: $\varrho(\vartheta) = 2 f \cdot \tan(\vartheta/2)$ 
\item Double Sphere:\newline $\varrho(\vartheta) = f\cdot \frac{\sin\vartheta}{\alpha\sqrt{\sin^2\vartheta + (\xi + \cos\vartheta)^2} + (1-\alpha)(\xi + \cos\vartheta)}$ \\
\end{enumerate}
The distortion coefficients $a_1, a_2, a_3, a_4$ are used to create the angle of incidence maps $(a_x,a_y)$, $c_x, c_y$ are used to compute the principal point coordinate maps $(cc_x,cc_y)$, and the camera's sensor dimensions (width $w$ and height $h$) are used to formulate the are utilized to formulate the normalized coordinate maps. We compute these maps using the camera intrinsic parameters and include them in our shared self-attention encoder.\par
\textbf{\textit{Centered Coordinates ($cc$)}} 
The SAN \emph{pairwise} and \emph{patchwise} operation modules receive the principal point position via the $cc_x$ and $cc_y$ coordinate channels, which are centred at $(0,0)$. We create $cc_x$ and $cc_y$ channels, as shown below:
\begin{equation}
\label{eq:cc}
    \resizebox{0.85\hsize}{!}{%
    $
    cc_x=\begin{pmatrix}
           \smash{0 -c_x}\\
           \smash{1 -c_x}\\
           \smash{\vdots} \\
           \smash{w-c_x}
    \end{pmatrix}\cdot
    \begin{pmatrix}
           1 \\
           1 \\
           \smash{\vdots} \\
           1
    \end{pmatrix}_{(h+1)\times 1}^{\intercal}
    cc_y=\begin{pmatrix}
           1 \\
           1 \\
           \smash{\vdots} \\
           1
    \end{pmatrix}_{(w+1)\times 1}\cdot\begin{pmatrix}
           \smash{0 -c_y}\\
           \smash{1 -c_y}\\
           \smash{\vdots} \\
           \smash{h - c_y}
    \end{pmatrix}^{\intercal}
    $
    }
\end{equation}
We concatenate $cc_x$ and $cc_y$ by resizing them with bilinear interpolation to match the input feature's size.\par
\textbf{\textit{Angle of Incidence Maps $(a_x,a_y)$}}
Using the focal length $f$ of the camera, the horizontal and vertical angle of incidence maps for the pinhole (Rectilinear) camera model are determined from the $cc$ maps.
\begin{equation}
    \label{eq:ff}
    a_{ch}[i,j] = \arctan\Big(\frac{cc_{ch}[i,j]}{f}\Big)
\end{equation}
where $ch$ represents either $x$ or $y$ (see Eq.~\ref{eq:cc}). By computing the inverse of the radial distortion functions $\varrho(\vartheta)$, angle of incidence maps for different fisheye camera models can be calculated. The angle of incidence $\vartheta$ is computed numerically for the polynomial model employed in this paper by computing the $4^\text{th}$ order polynomial roots of $\varrho = \sqrt{x_I^2 + y_I^2} = a_1 \vartheta + a_2 \vartheta^2 + a_3 \vartheta^3 + a_4 \vartheta^4$. To improve training efficiency, we store the pre-calculated roots in a lookup table for all pixel coordinates and generate the $a_x$ and $a_y$ maps by setting $x_I = cc_x[i,j], y_I = 0$ and $x_I = 0, y_I = cc_y[i,j]$, respectively. A lookup table is not necessary for UCM, eUCM, and Double Sphere projection models as they can be inverted analytically.\par
\textbf{\textit{Normalized Coordinates ($nc$)}} Following~\cite{liu2018intriguing, Facil2019}, we additionally introduce two channels of normalized coordinates, whose values span linearly with respect to the image coordinates between $-1$ and $1$. The channels are unaffected by camera sensor properties and characterize the spatial extent of the content in feature space in each direction (\eg, a $\hat{x}$ channel value close to $1$ indicates that the feature vector at this location is near the right border).\par
\subsection{\textbf{Semantically-Guided Distance Decoder}}
To better guide the distance features, we adapt the pixel-adaptive convolutions from \cite{su2019pixel, Guizilini2020}, thereby distilling the knowledge of the extracted features from the segmentation branch into the distance decoder. This follows the intuition that the distance estimation can profit from the location-specific knowledge encoded in the semantic features, which in return also breaks up the convolutions' spatial invariance. While other semantic guidance approaches relied on a pretrained segmentation network, \cite{Casser2019, Guizilini2020}, our semantic segmentation branch is learned simultaneously with the distance estimation, thereby introducing a more favorable one-stage training.\par
To implement the pixel-adaptive convolutions, we rely on the multi-level extraction of features from the segmentation decoder, as shown in Fig.~\ref{fig:model_arch}. Accordingly, the input $x$ is processed as follows:
\begin{equation}
x_{ij}' = \sum_{ab \in \mathcal{N}_k(i, j)} K(F_{ij},F_{ab}) W [r_{a-i,b-j}]x_{ab} + B
\label{eq:pixel-convolutions}
\end{equation}
where $r_{a-i,b-j}$ represents the distance between pixel locations $ij$ and $ab$ and $\mathcal{N}_k(i, j)$ defines a neighborhood window of size $k \times k$ centered around the location $ij$. In Eq.~\ref{eq:pixel-convolutions}, the elements of $x$ which are covered by the neighborhood window $\mathcal{N}_k(i, j)$ make up the input to the convolution, characterized by a kernel function $K$, weights $W$, and a bias $B\in \mathbb{R}^1$. The kernel function is necessary to compute the correlation between the semantic features $F \in \mathbb{R}^D$ extracted by the segmentation decoder. Following \cite{Guizilini2020}, we define the kernel function as:
\begin{equation}
K(F_{ij},F_{ab}) = \exp\left(-\frac{1}{2} (F_{ij} - F_{ab})^T \Sigma_{ijab}^{-1} (F_{ij} - F_{ab})\right)
\end{equation}
where $\Sigma_{ijab}$ is the covariance matrix between the features $F_{ij}$ and $F_{ab}$. This matrix is chosen to be diagonal, which means that it can be expressed in terms of a learnable parameter $\sigma$ for each convolutional filter as $\sigma^2 \cdot\mathbf{1}^D$.\par

%% file: include/implementation_details.tex
\section{Implementation Details}
\label{sec:implementation_details}

To minimise the training objective function, we use Pytorch to implement the models described in Section~\ref{sec:network_details} and the Ranger (RAdam~\cite{liu2019variance} + LookAhead~\cite{zhang2019lookahead}) optimizer. RAdam uses an active rectifier to adjust Adam's~\cite{kingma2014adam} adaptive momentum, resulting in an automated warm-up that is tailored to the dataset of interest. Meanwhile, LookAhead introduces a set of ``slow weights'', which are updated based on the previous weight updates of the normal optimizer (RAdam).
The combination of both methods is highly synergistic, resulting in improved results. We train the model for 10 epochs because we now have 4x the data compared to training only the front camera images on a 24GB Titan RTX with a batch size of 20. We use a learning rate of ${4 \times {10}^{-4}}$, which we reduce to ${{10}^{-5}}$ after 7 epochs for another 3 epochs. The distance decoder's sigmoid output $\sigma$ is converted to distance using $D = {m \cdot \sigma + n}$. Depth is calculated for the pinhole model as $D = 1 / ({m \cdot \sigma + n})$, where $n$ and $m$ are constants chosen such that $D$ is constrained between $0.1$ and $100$. As described in Section~\ref{sec:camera-geometry-tensor} (also cf.~Fig.~\ref{fig:model_arch}), we shuffle all images from the surround-view cameras with diverse viewpoints and present them to the distance and pose models along with their corresponding intrinsics to build the camera geometry tensor $C_t$. We limited the length of the sequences to three. The lengths of the sequences considered for pose estimation results are 2, 3, and 5. The semantic segmentation is trained supervised by applying the cross-entropy loss and optimized jointly with the distance estimation (see Fig.~\ref{fig:mtl_pipeline}).\par
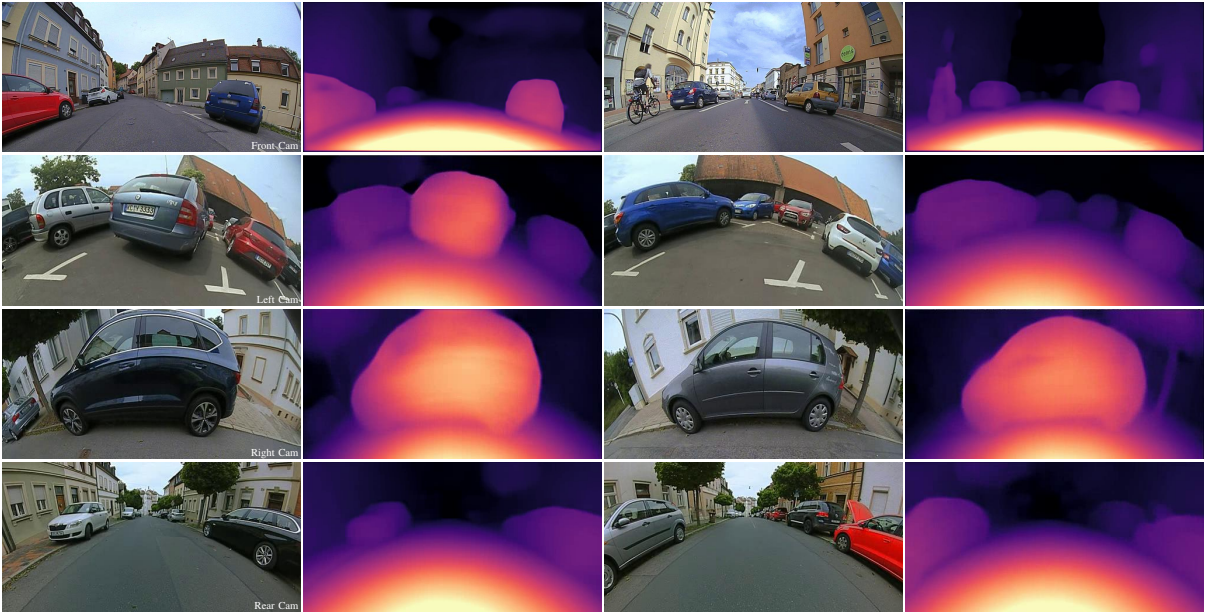
\begin{figure*}[!t]
\captionsetup{belowskip=-8pt, font= small, singlelinecheck=false}
\centering
  \resizebox{0.9\textwidth}{!}{\input{img/multi_camera/multi_cam}}
\caption{{\bf Distance estimation results} of the same network evaluated on four different fisheye cameras of a surround-view camera system. One can see that our SVDistNet model generalizes well across different viewing angles and consistently produces high-quality distance outputs.}
  \label{fig:multi-cam}
\end{figure*}
\textbf{\textit{Comparison to Convolutional Networks}}
Pairwise models match or outperform convolutional baselines while requiring comparable or lower parameters and FLOP budgets. In terms of computational complexity, the patchwise models outperform CNN's. For example, the patchwise SAN10 with $11.8$M params and $1.9$G FLOPS outperforms ResNet50 with $25.6$M params and $4.1$G FLOPS, while having a 54\% lower parameter and 44\% lower FLOP count. SAN10-patch models' parameter count is almost equivalent to ResNet18 with $11.7$M params and $1.8$G FLOPS, whereas SAN15-patch with $20.5$M params and $3.3$G FLOPS is equivalent to ResNet50's parameter count. The SAN10-pairwise with $10.5$M params and $2.2$G FLOPS outperforms ResNet18 with a 9\% lower parameter count and 22\% higher FLOP count. By performing a single-scale image depth prediction rather than the multi-scale in our previous works~\cite{kumar2020fisheyedistancenet, kumar2020unrectdepthnet}, we could leverage the use of a more robust loss function over \lone to reduce training times on SAN10.\par

%% file: img/multi_camera/multi_cam.tex
\newcommand{\imlabel}[2]{\includegraphics[width=.49\linewidth]{#1}%
\raisebox{2pt}{\makebox[-2pt][r]{\small #2}}}

\scalebox{1}[1]{
\begin{tabular}{@{\hskip 1.5mm}c@{\hskip 1.5mm}c @{\hskip 1.5mm}c@{\hskip 1.5mm}c}
\centering
    \imlabel{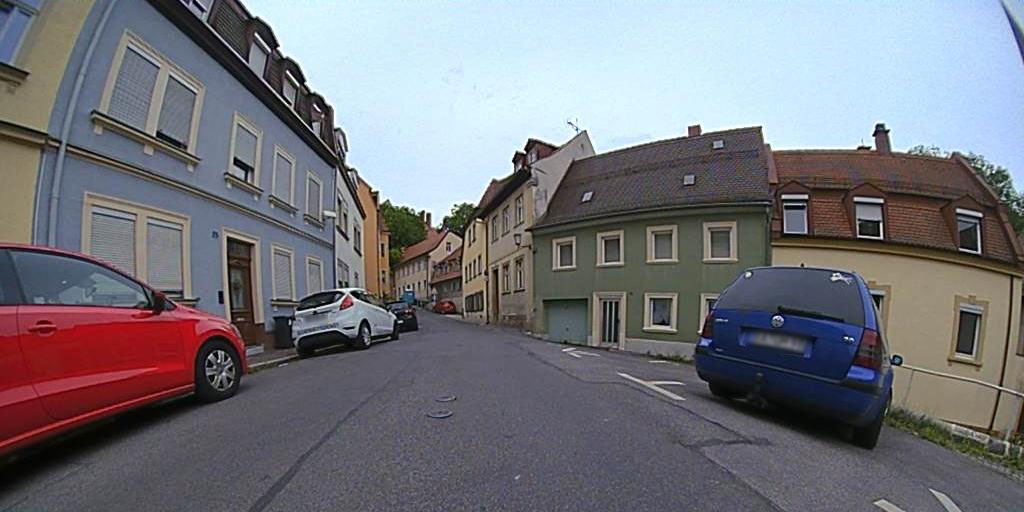}
    {\hspace{.24\textwidth}\textcolor{white}{Front Cam}} &
    \imlabel{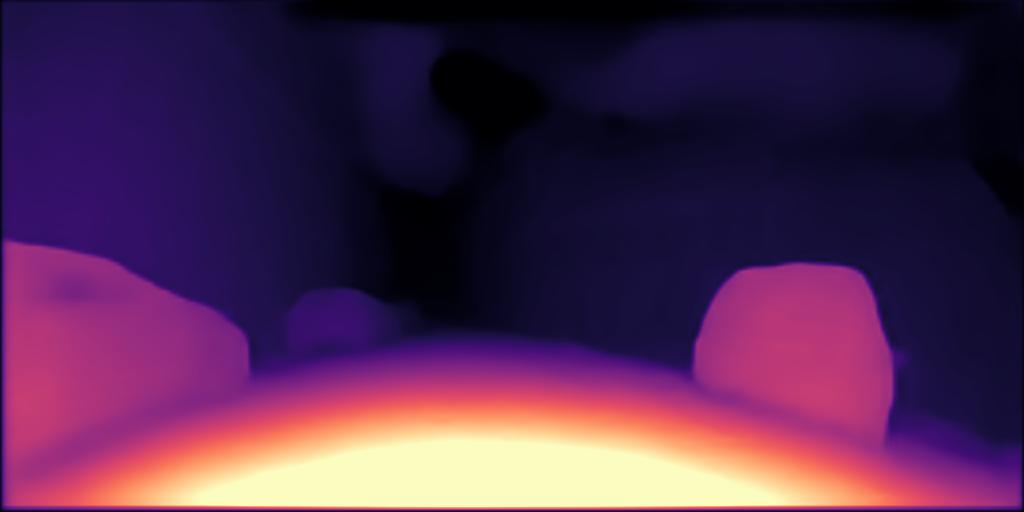}
    {} &
    \imlabel{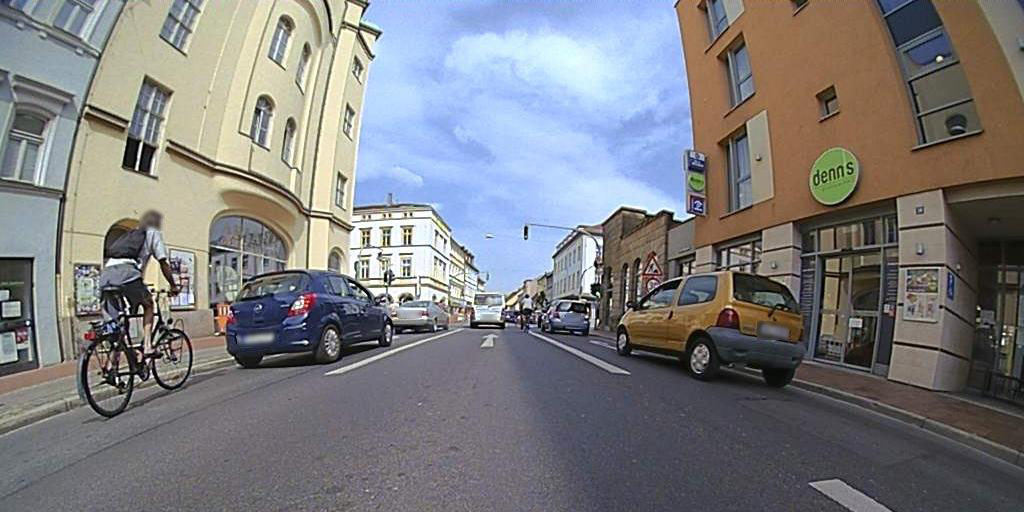}
    {} &
    \imlabel{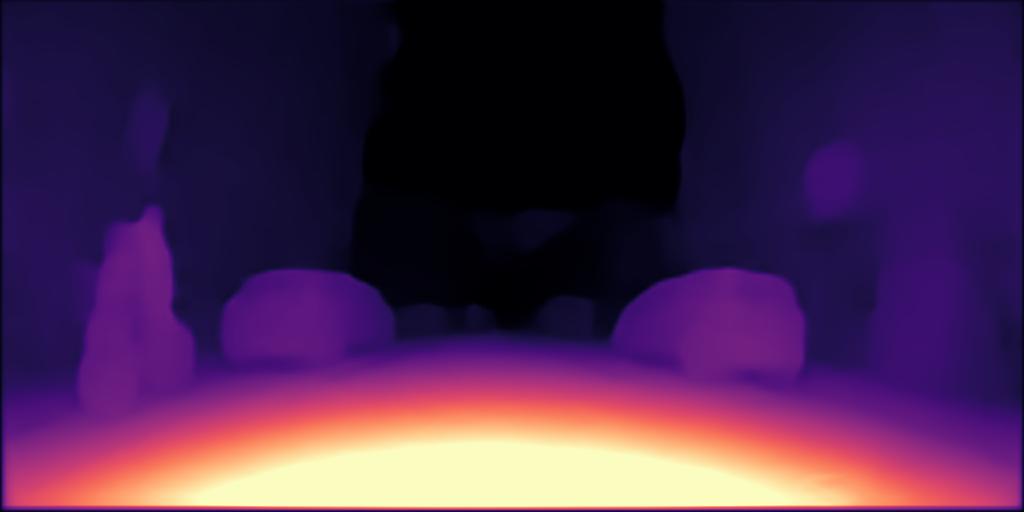}
    {} \\

    \imlabel{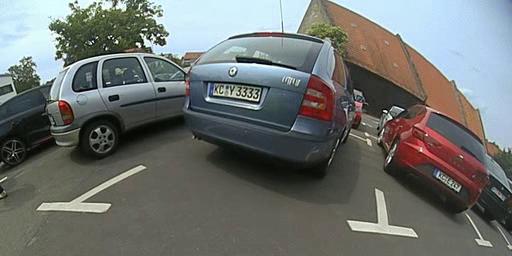}
    {\hspace{.24\textwidth}\textcolor{white}{Left Cam}} &
    \imlabel{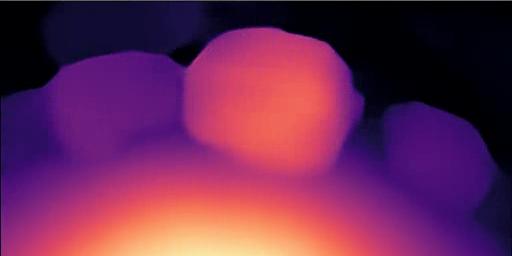}
    {} &
    \imlabel{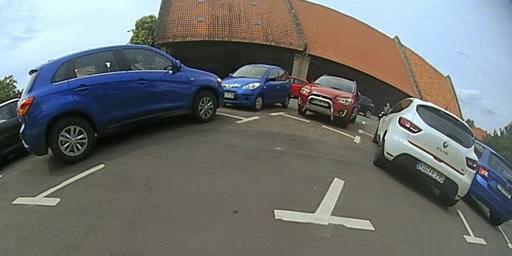}
    {} &
    \imlabel{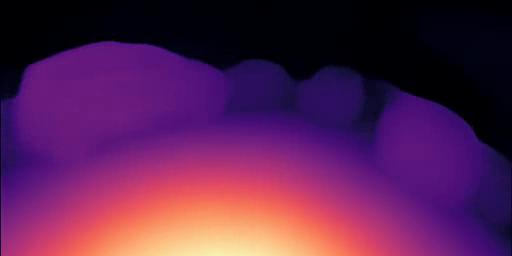}
    {} \\
    
    \imlabel{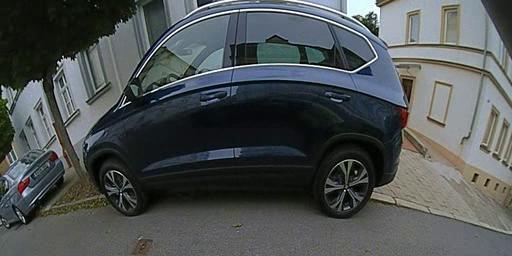}
    {\hspace{.24\textwidth}\textcolor{white}{Right Cam}} &
    \imlabel{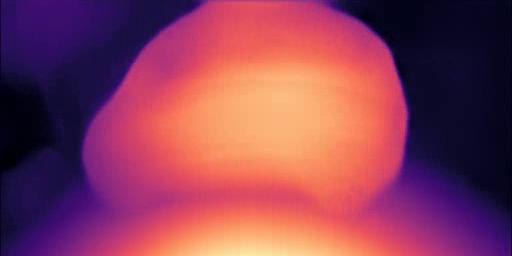}
    {} &
    \imlabel{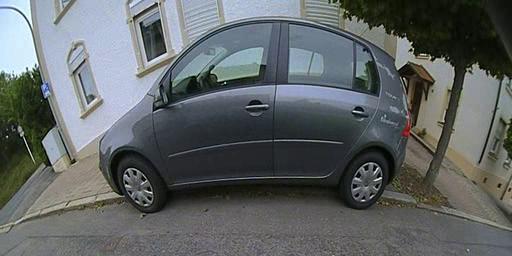}
    {} &
    \imlabel{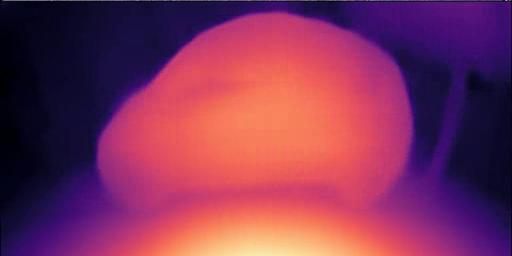}
    {} \\
    
    \imlabel{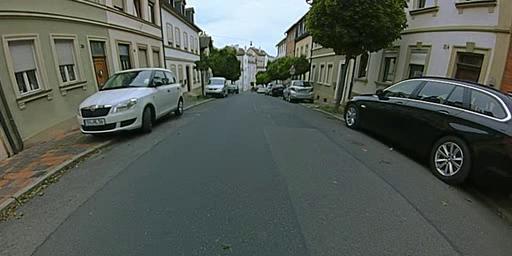}
    {\hspace{.24\textwidth}\textcolor{white}{Rear Cam}} &
    \imlabel{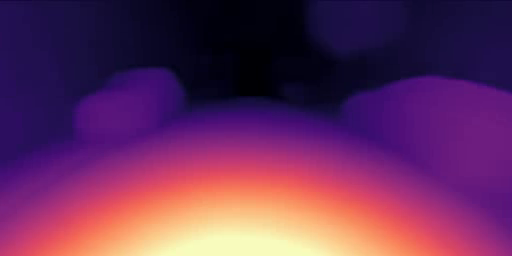}
    {} &
    \imlabel{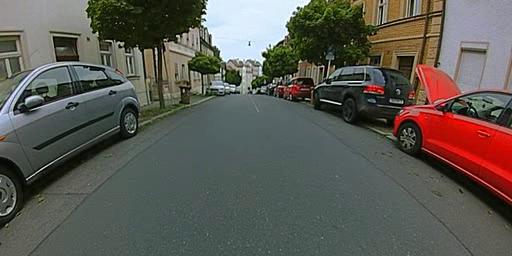}
    {} &
    \imlabel{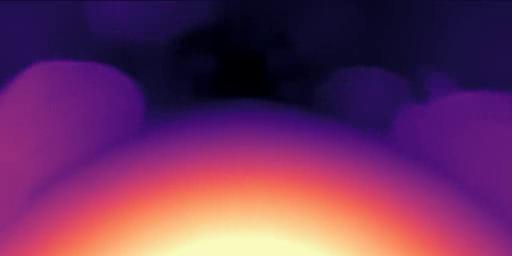}
    {} \\
\end{tabular}
}

%% file: include/experiments.tex
\section{Experimental Evaluation} 
\label{sec:experiments}

\subsection{\textbf{Effect of Multi-Task Learning}}

\input{results/mtl}
\input{results/features_ablation}
In Table~\ref{tab:mtl}, we compare our SVDistnet with previous work~\cite{kumar2020syndistnet} on the Woodscape dataset \cite{yogamani2019woodscape}. Specifically, we evaluate our model using the patchwise self-attention encoder. First, we create single-task baselines using the various methods mentioned above, and then we create an important shared encoder-based multi-task learning (MTL) baseline for these tasks. Due to the use of an improved self-attention encoder and the camera geometry tensor, the MTL results and single-task benchmark for distance estimation outperform SynDistNet~\cite{kumar2020syndistnet}. However, we observe only minimal gain for the semantic segmentation task. In the final experiment, we include the synergy between the distance and segmentation decoders. For these diverse tasks, we observe that the content and channel adaptive self-attention encoder features can be learned jointly, wherein the captured semantic features, used along with pixel-adaptive convolutions, guide the distance decoder to capture better geometric features. We use ablation experiments to further dissect these findings.\par
\subsection{\textbf{Ablation Study on the Single Contributions}}

For our ablation analysis, we consider two self-attention encoder variants: pairwise and patchwise. Table~\ref{table:table_ablation} displays the distance estimation results for these variants. We show the impact of certain contributions and their importance in our SVDistNet framework compared to our previous work SynDistNet~\cite{kumar2020syndistnet}. As a start, we replace the \lone loss with the general robust loss function and test it on the self-attention encoder's patchwise variant, which results in significant accuracy improvements. To filter all the dynamic objects that contaminate the reconstruction loss, we use the dynamic object mask obtained by projecting semantic segmentation predictions as described in  Section~\ref{sec:dynamic-object-mask}. Furthermore, this contribution has the potential to solve the infinite distance problem.\par

We see a significant increase in accuracy after adding the camera geometry tensor to this setting. The training on multiple cameras with varying intrinsics and viewing angles adds an additional regularization. The network's aforementioned training strategy is thereby more camera-independent and leads to a better generalization to images captured with varying cameras. To improve our multi-task setup even further, we use a distance and segmentation decoder synergy. We apply semantic features from the supervised task to the distance decoder's geometric features while still training the distance estimation in a self-supervised manner. In this case, we can remove the dynamic object mask and still see improvements.\par

By adding the camera geometry tensor to this setting, we can improve the metric results even further and potentially outperform the best setting in SynDistNet~\cite{kumar2020syndistnet}. Using these critical features, we create an experiment similar to SynDistNet's final setting, which combines all of the listed features except the camera geometry tensor. We could attain better results than SynDistNet's ResNet50 results. It is important to note that ResNet50 is comparable to the SAN19-pair/patch encoder. However, in terms of computational complexity, we were able to outperform ResNet50 with the SAN10-patch encoder (cf. Section \ref{sec:implementation_details}).\par

By introducing our camera geometry tensor, we completed our SVDistNet model for the surround view-camera framework. We also performed a few vital evaluations using the pairwise self-attention encoder head. We were unable to achieve the same level of accuracy as the patchwise self-attention encoder. The patchwise self-attention module outperforms convolution, while the pairwise self-attention module equals or outperforms the convolutional equivalents. We use a higher-performing SAN19-patch encoder network to perform the same vital ablation of our contributions. Finally, with all of our additions, we outperform all previous works on the Woodscape dataset~\cite{kumar2020fisheyedistancenet, kumar2020unrectdepthnet, kumar2020syndistnet}.\par
\input{results/multi_camera}
\subsection{\textbf{Generalization Across Different Cameras}}

The generalization across different cameras from the surround-view setup is depicted in Table~\ref{table:multi-camera-ablation} using the camera geometry tensor as described in Section~\ref{sec:camera-geometry-tensor}. Each camera's metrics significantly improve as sequences from different cameras aid in generalization during the model's training phase. For example, the front camera's distance estimates profit as the side cameras steer the network to generalize close and overlapping objects. Because of the use of $C_t$, our network does not overfit to a specific camera intrinsic. It adapts to any changes detected by a family of unseen cameras deployed with a pre-calibrated camera setup. This leads to better estimates and generalization of new cameras and the ability to train on images from different cameras.\par
\input{results/quantitative}
\subsection{\textbf{State-of-the-Art Comparison on KITTI}}

To facilitate comparisons with previous methods, we also train our distance estimation method on the KITTI Eigen split~\cite{geiger2013vision, Eigen_14} in the classical depth estimation setting. First, we train and evaluate on depth maps generated from LiDAR point clouds, where Table~\ref{tab:results} shows that with our contributions, textit{we outperform all previous methods}.\par

We obtain a significant improvement using the online refinement method from~\cite{Casser2019}, and our results remain superior to previous methods. Also, when training and evaluating the improved KITTI labels for depth estimation, we show a significant improvement. Simultaneously, our network architecture's complexity is still comparable to the one of a ResNet18 in terms of computational complexity. 
We also use the general camera tensor $C_t$ as described in Section~\ref{sec:camera-geometry-tensor} in our model, wherein instead of the angle of incidence maps, we employ the maps generated using Eq.~\ref{eq:ff} for pinhole cameras.\par

%% file: results/mtl.tex
\begin{table}[t]
\captionsetup{belowskip=-12pt, skip=4pt, font= small, singlelinecheck=false}
  \centering
    \begin{adjustbox}{width=\columnwidth}
    \setlength{\tabcolsep}{0.2em}
      \begin{tabular}{lcccccc}
      \toprule
      \multicolumn{1}{l}{\textbf{Method}} & 
      \multicolumn{1}{c}{\textit{\cellcolor[HTML]{00b050}Seg}} &
      \multicolumn{1}{c}{\cellcolor[HTML]{00b0f0}\textit{Dist.}} &
      \multicolumn{1}{c}{\cellcolor[HTML]{ab9ac0}\textit{MTL}} &
      \multicolumn{1}{c}{\cellcolor[HTML]{e5b9b5}\begin{tabular}[c]{@{}c@{}} \textit{MTL} \\ \textit{(Synergy)} \end{tabular}} &
      \multicolumn{1}{c}{\cellcolor[HTML]{00b050}\begin{tabular}[c]{@{}c@{}} \textit{Seg.} \\ \textit{(mIOU)} \end{tabular}} 
      & \multicolumn{1}{c}{\cellcolor[HTML]{00b0f0}\begin{tabular}[c]{@{}c@{}} \textit{Distance} \\ \textit{(RMSE)} \end{tabular}} \\ 
      \toprule
      \multirow{4}{*}{\begin{tabular}[c]{@{}c@{}} SynDistNet~\cite{kumar2020syndistnet} \end{tabular}}
      & \ch & \xm & \xm & \xm & 76.8          & \xm   \\ 
      & \xm & \ch & \xm & \xm & \xm           & 2.316 \\
      & \ch & \ch & \ch & \xm & 78.3          & 2.128 \\ 
      & \ch & \ch & \xm & \ch & \textbf{81.5} & \textbf{1.714} \\
      \midrule
      \multirow{4}{*}{\begin{tabular}[c]{@{}c@{}} SVDistNet \\ (SAN10-patch) \end{tabular}}
      & \ch & \xm & \xm & \xm & 77.1          &  \xm   \\ 
      & \xm & \ch & \xm & \xm & \xm           & 2.153 \\
      & \ch & \ch & \ch & \xm & 78.6          & 1.861 \\ 
      & \ch & \ch & \xm & \ch & \textbf{82.3} & \textbf{1.532} \\
      \bottomrule
    \end{tabular}
\end{adjustbox}
\caption{\textbf{Effect of our multi-task training approach} on semantic segmentation and distance estimation when using a ResNet 18 backbone for SynDistNet~\cite{kumar2020syndistnet} and self-attention network SAN-10 for SVDistNet. We observe that both outputs improve through the multi-task training and mainly the distance estimation performance profits from the synergized semantic guidance.}
\label{tab:mtl}
\end{table}

%% file: results/features_ablation.tex
\begin{table*}[!t]
\captionsetup{belowskip=-10pt, skip=2pt, font= small, singlelinecheck=false}
  \centering{
  \small
  \setlength{\tabcolsep}{0.18em}
  \begin{tabular}{lcccccccccccc}
    \toprule
        \multirow{2}{*}{\textbf{Network}} & 
        \multirow{2}{*}{\emph{RL}} &
        \multirow{2}{*}{\emph{Self-Attn}} & 
        \multirow{2}{*}{\emph{SEM}} &
        \multirow{2}{*}{\emph{Mask}} &
        \multirow{2}{*}{\emph{CGT}} &
        \cellcolor[HTML]{7d9ebf}Abs Rel & \cellcolor[HTML]{7d9ebf}Sq Rel & \cellcolor[HTML]{7d9ebf}RMSE & \cellcolor[HTML]{7d9ebf}RMSE$_{log}$ & \cellcolor[HTML]{e8715b}$\delta<1.25$ & \cellcolor[HTML]{e8715b}$\delta<1.25^2$ & \cellcolor[HTML]{e8715b}$\delta<1.25^3$ \\
        \cmidrule(lr){7-10} \cmidrule(lr){11-13}
         &  &   &  &  &  & \multicolumn{4}{c}{\cellcolor[HTML]{7d9ebf}lower is better} & \multicolumn{3}{c}{\cellcolor[HTML]{e8715b}higher is better} \\
    \toprule
        FisheyeDistanceNet~\cite{kumar2020fisheyedistancenet} 
        & \xm & \xm & \xm & \xm & \xm & 0.152 & 0.768 & 2.723 & 0.210 & 0.812 & 0.954 & 0.974 \\
    \midrule
        \multirow{6}{*}{SynDistNet(ResNet18)~\cite{kumar2020syndistnet}}
        & \ch & \xm & \xm & \xm & \xm & 0.142 & 0.537 & 2.316 & 0.179 & 0.878 & 0.971 & 0.985 \\
        & \ch & \xm & \xm & \ch & \xm & 0.133 & 0.491 & 2.264 & 0.168 & 0.868 & 0.976 & 0.988 \\
        & \ch & \ch & \xm & \ch & \xm & 0.121 & 0.429 & 2.128 & 0.155 & 0.875 & 0.980 & 0.990 \\
        & \ch & \ch & \ch & \xm & \xm & 0.105 & 0.396 & 1.976 & 0.143 & 0.878 & 0.982 & 0.992 \\
        & \ch & \ch & \ch & \ch & \xm & \textbf{0.076} & \textbf{0.368} & \textbf{1.714} & \textbf{0.127} & \textbf{0.891} & \textbf{0.988} & \textbf{0.994} \\
    \midrule
        \multirow{6}{*}{SynDistNet(ResNet50)~\cite{kumar2020syndistnet}}
        & \ch & \xm & \xm & \xm & \xm & 0.138 & 0.540 & 2.279 & 0.177 & 0.880 & 0.973 & 0.986 \\
        & \ch & \xm & \xm & \ch & \xm & 0.127 & 0.485 & 2.204 & 0.166 & 0.881 & 0.975 & 0.989 \\
        & \ch & \ch & \xm & \ch & \xm & 0.115 & 0.413 & 2.028 & 0.148 & 0.876 & 0.983 & 0.992 \\
        & \ch & \ch & \ch & \xm & \xm & 0.102 & 0.387 & 1.856 & 0.135 & 0.884 & 0.985 & 0.994 \\
        & \ch & \ch & \ch & \ch & \xm & \textbf{0.068} & \textbf{0.352} & \textbf{1.668} & \textbf{0.121} & \textbf{0.895} & \textbf{0.990} & \textbf{0.996} \\
    \midrule
        \multirow{6}{*}{SVDistNet (SAN10-patch)}
        & \ch & \ch & \xm & \xm & \xm & 0.128 & 0.469 & 2.153 & 0.164 & 0.875 & 0.974 & 0.986 \\
        & \ch & \ch & \xm & \ch & \xm & 0.114 & 0.413 & 2.022 & 0.149 & 0.878 & 0.982 & 0.990 \\
        & \ch & \ch & \xm & \ch & \ch & 0.101 & 0.378 & 1.861 & 0.133 & 0.884 & 0.984 & 0.991 \\
        & \ch & \ch & \ch & \xm & \xm & 0.094 & 0.345 & 1.789 & 0.128 & 0.887 & 0.985 & 0.992 \\
        & \ch & \ch & \ch & \xm & \ch & 0.082 & 0.316 & 1.682 & 0.119 & 0.890 & 0.987 & 0.993 \\
        & \ch & \ch & \ch & \ch & \xm & 0.074 & 0.343 & 1.641 & 0.112 & 0.892 & 0.985 & 0.994 \\
        & \ch & \ch & \ch & \ch & \ch &\textbf{0.057} & \textbf{0.315} & \textbf{1.532} & \textbf{0.108} & 
\textbf{0.896} & \textbf{0.986} & \textbf{0.996} \\
    \midrule
        \multirow{2}{*}{SVDistNet (SAN10-pair)}
        & \ch & \ch & \xm & \ch & \ch & 0.121 & 0.457 & 2.115 & 0.152 & 0.879 & 0.979 & 0.985 \\
        & \ch & \ch & \ch & \xm & \ch & 0.103 & 0.385 & 1.882 & 0.141 & 0.882 & 0.983 & 0.990 \\
        & \ch & \ch & \ch & \ch & \ch & \textbf{0.081} & \textbf{0.365} & \textbf{1.710} & \textbf{0.128} & \textbf{0.890} & \textbf{0.985} & \textbf{0.994} \\
    \midrule
        \multirow{6}{*}{SVDistNet (SAN19-patch)}
        & \ch & \ch & \xm & \xm & \xm & 0.121 & 0.437 & 2.127 & 0.153 & 0.878 & 0.976 & 0.989 \\
        & \ch & \ch & \xm & \ch & \xm & 0.109 & 0.408 & 2.006 & 0.145 & 0.880 & 0.982 & 0.992 \\
        & \ch & \ch & \xm & \ch & \ch & 0.098 & 0.372 & 1.849 & 0.138 & 0.884 & 0.983 & 0.991 \\
        & \ch & \ch & \ch & \xm & \xm & 0.091 & 0.351 & 1.773 & 0.129 & 0.886 & 0.986 & 0.993 \\
        & \ch & \ch & \ch & \xm & \ch & 0.070 & 0.305 & 1.669 & 0.108 & 0.893 & 0.986 & 0.994 \\
        & \ch & \ch & \ch & \ch & \xm & 0.067 & 0.296 & 1.578 & 0.106 & 0.895 & 0.985 & 0.994 \\
        & \ch & \ch & \ch & \ch & \ch &\textbf{0.048} & \textbf{0.277} & \textbf{1.486} & \textbf{0.086} & \textbf{0.901} & \textbf{0.991} & \textbf{0.996} \\
    \midrule
        \multirow{2}{*}{SVDistNet (SAN19-pair)}
        & \ch & \ch & \xm & \ch & \ch & 0.116 & 0.461 & 2.097 & 0.154 & 0.881 & 0.982 & 0.988 \\
        & \ch & \ch & \ch & \xm & \ch & 0.096 & 0.371 & 1.846 & 0.147 & 0.884 & 0.985 & 0.991 \\
        & \ch & \ch & \ch & \ch & \ch & \textbf{0.074} & \textbf{0.331} & \textbf{1.624} & \textbf{0.101} & \textbf{0.891} & \textbf{0.986} & \textbf{0.994} \\
        \bottomrule
  \end{tabular}
  }
\caption{\textbf{Ablation study on the effect of our contributions} up to our final SVDistNet model on the Fisheye Woodscape dataset \cite{yogamani2019woodscape}. We cap the distance estimates to 40m. From our distance estimation baseline \cite{kumar2020fisheyedistancenet}, we incrementally add up the robust loss (RL), self-attention layers encoder heads (Self-Attn), semantic guidance in the decoder (SEM), dynamic object masking (Mask), and camera geometry tensor (CGT). We showcase our improvements for various network architectures and, in particular, show the superiority of our SVDistNet model over the SynDistNet model as well as the positive effect of using camera geometry tensor $C_t$.}
\label{table:table_ablation}
\end{table*}

%% file: results/multi_camera.tex
\begin{table}[t]
\captionsetup{belowskip=-8pt, font= small, singlelinecheck=false}
    \centering
    \small
    \begin{adjustbox}{width=\columnwidth}
    \setlength{\tabcolsep}{0.2em}
      \begin{tabular}{lcccccccc}
	  \toprule
      \multirow{2}{*}{\textbf{Cams}} & 
      \multirow{2}{*}{\emph{CGT}} & 
      \cellcolor[HTML]{7d9ebf}Abs Rel & \cellcolor[HTML]{7d9ebf}Sq Rel & \cellcolor[HTML]{7d9ebf}RMSE & \cellcolor[HTML]{7d9ebf}RMSE$_{log}$ & \cellcolor[HTML]{e8715b}$\delta<1.25$ & \cellcolor[HTML]{e8715b}$\delta<1.25^2$ & \cellcolor[HTML]{e8715b}$\delta<1.25^3$ \\
      \cmidrule(lr){3-6} \cmidrule(lr){7-9}
      & & \multicolumn{4}{c}{\cellcolor[HTML]{7d9ebf}lower is better} & \multicolumn{3}{c}{\cellcolor[HTML]{e8715b}higher is better} \\
      \toprule
      \multirow{2}{*}{\begin{tabular}[c]{@{}c@{}} Front \end{tabular}}             
      & \xm & 0.074 & 0.343 & 1.641 & 0.112 & 0.892 & 0.985 & 0.994 \\
      & \ch & 0.057 & 0.315 & 1.532 & 0.108 & 0.896 & 0.987 & 0.996 \\
      \midrule
      \multirow{2}{*}{\begin{tabular}[c]{@{}c@{}} Rear \end{tabular}}              
      & \xm & 0.089 & 0.358 & 1.657 & 0.131 & 0.888 & 0.981 & 0.988 \\
      & \ch & 0.065 & 0.337 & 1.579 & 0.123 & 0.891 & 0.983 & 0.992 \\
      \midrule
      \multirow{2}{*}{\begin{tabular}[c]{@{}c@{}} Left \end{tabular}}  
      & \xm & 0.102 & 0.398 & 1.874 & 0.126 & 0.886 & 0.980 & 0.983 \\
      & \ch & 0.091 & 0.382 & 1.781 & 0.114 & 0.889 & 0.985 & 0.990 \\
      \midrule
      \multirow{2}{*}{\begin{tabular}[c]{@{}c@{}} Right \end{tabular}}  
      & \xm & 0.105 & 0.406 & 1.889 & 0.135 & 0.882 & 0.980 & 0.981 \\
      & \ch & 0.093 & 0.391 & 1.796 & 0.120 & 0.887 & 0.983 & 0.986 \\
      \bottomrule
  \end{tabular}
\end{adjustbox}
\caption{\textbf{Ablation study on multiple cameras} concerning the usage of Camera Geometry Tensor using WoodScape~\cite{yogamani2019woodscape}.}
\label{table:multi-camera-ablation}
\end{table}

%% file: results/quantitative.tex
\begin{table}[t!]
\captionsetup{belowskip=-14pt, font= small, singlelinecheck=false}
\renewcommand{\arraystretch}{0.87}
\centering
{
\small
\setlength{\tabcolsep}{0.18em}
\begin{adjustbox}{width=\columnwidth}
\begin{tabular}{c|lccccccc}
\toprule
& \multirow{2}{*}{\textbf{Method}}
& \cellcolor[HTML]{7d9ebf}Abs$_{rel}$ 
& \cellcolor[HTML]{7d9ebf}Sq$_{rel}$ 
& \cellcolor[HTML]{7d9ebf}RMSE 
& \cellcolor[HTML]{7d9ebf}RMSE$_{log}$ 
& \cellcolor[HTML]{e8715b}$\delta_1$ 
& \cellcolor[HTML]{e8715b}$\delta_2$ 
& \cellcolor[HTML]{e8715b}$\delta_3$ \\
\cmidrule(lr){3-6} \cmidrule(lr){7-9}
&  & \multicolumn{4}{c}{\cellcolor[HTML]{7d9ebf}lower is better} & \multicolumn{3}{c}{\cellcolor[HTML]{e8715b}higher is better} \\
\toprule
\parbox[t]{2.5mm}{\multirow{10}{*}{\rotatebox[origin=c]{90}{Original~\cite{Eigen_14}}}}
& Monodepth2~\cite{Godard2019}                & 0.115 & 0.903 & 4.863 & 0.193 & 0.877 & 0.959 & 0.981 \\
& PackNet-SfM~\cite{Guizilini2020a}           & 0.111 & 0.829 & 4.788 & 0.199 & 0.864 & 0.954 & 0.980 \\
& FisheyeDistanceNet~\cite{kumar2020fisheyedistancenet} & 0.117 & 0.867 & 4.739 & 0.190 & 0.869 & 0.960 & 0.982 \\
& UnRectDepthNet~\cite{kumar2020unrectdepthnet} & 0.107 & 0.721 & 4.564 & 0.178 & 0.894 & 0.971 & 0.986 \\
& SynDistNet~\cite{kumar2020syndistnet}      & 0.109 & 0.718 & 4.516 & 0.180 & 0.896 & 0.973 & 0.986 \\
& Shu \etal~\cite{shu2020featdepth}          & 0.104 & 0.729 & 4.481 & 0.179 & 0.893 & 0.965 & 0.984 \\
& SVDistNet                                  & \textbf{0.102} & \textbf{0.706} & \textbf{4.459} & \textbf{0.172} & \textbf{0.908} & \textbf{0.974} & \textbf{0.986} \\
\cmidrule{2-9}
& Struct2Depth${^*}$~\cite{Casser2019}       & 0.109 & 0.825 & 4.750 & 0.187 &0.874 & 0.958 & 0.983 \\
& GLNet${^*}$~\cite{Chen2019b}               & 0.099 & 0.796 & 4.743 & 0.186 &0.884 & 0.955 & 0.979 \\
& Shu${^*}$ \etal~\cite{shu2020featdepth}    & 0.088 & 0.712 & 4.137 & \textbf{0.169} & 0.915 & 0.965 & 0.982 \\
& SVDistNet${^*}$                            & \textbf{0.086} & \textbf{0.701} & \textbf{4.118} & 0.170 & \textbf{0.919} & \textbf{0.976} & \textbf{0.985} \\
\midrule
\parbox[t]{2.5mm}{\multirow{6}{*}{\rotatebox[origin=c]{90}{Improved~\cite{uhrig2017sparsity}}}}
& Monodepth2~\cite{Godard2019}              & 0.090 & 0.545 & 3.942 & 0.137 & 0.914 & 0.983 & 0.995 \\
& PackNet-SfM~\cite{Guizilini2020a}         & 0.078 & 0.420 & 3.485 & 0.121 & 0.931 & 0.986 & 0.996 \\
& UnRectDepthNet~\cite{kumar2020unrectdepthnet} & 0.081 & 0.414 & 3.412 & 0.117 & 0.926 & 0.987 & 0.996 \\
& SynDistNet~\cite{kumar2020syndistnet}     & 0.076 & 0.412 & 3.406 & 0.115 & 0.931 & 0.988 & 0.996 \\
& SVDistNet                                 & \textbf{0.071} & \textbf{0.405} & \textbf{3.345} & \textbf{0.106} & \textbf{0.934} & \textbf{0.988} & \textbf{0.996} \\
& SVDistNet${^*}$                           & \textbf{0.059} & \textbf{0.392} & \textbf{3.206} & \textbf{0.097} & \textbf{0.935} & \textbf{0.989} & \textbf{0.995} \\
\bottomrule
\end{tabular}
\end{adjustbox}
}
\caption{\textbf{Evaluation of the KITTI Eigen split} in comparison to other self-supervised monocular depth estimation methods. We limit depths to 80\,m in accordance with best practices. We also evaluate using the \textit{Original} depth maps generated from raw point clouds as proposed by \cite{Eigen_14} and \textit{Improved} annotated depth maps as introduced by \cite{uhrig2017sparsity}. M indicates that sequences are trained on using the monocular approach. The method${^*}$ indicates the online refinement technique~\cite{Casser2019}, in which the model is trained during inference.}
\label{tab:results}
\end{table}

%% file: include/biographies.tex
\vspace{-3cm}
\begin{IEEEbiography}
[{\includegraphics[width=1.15in,height=1.5in,clip,keepaspectratio]{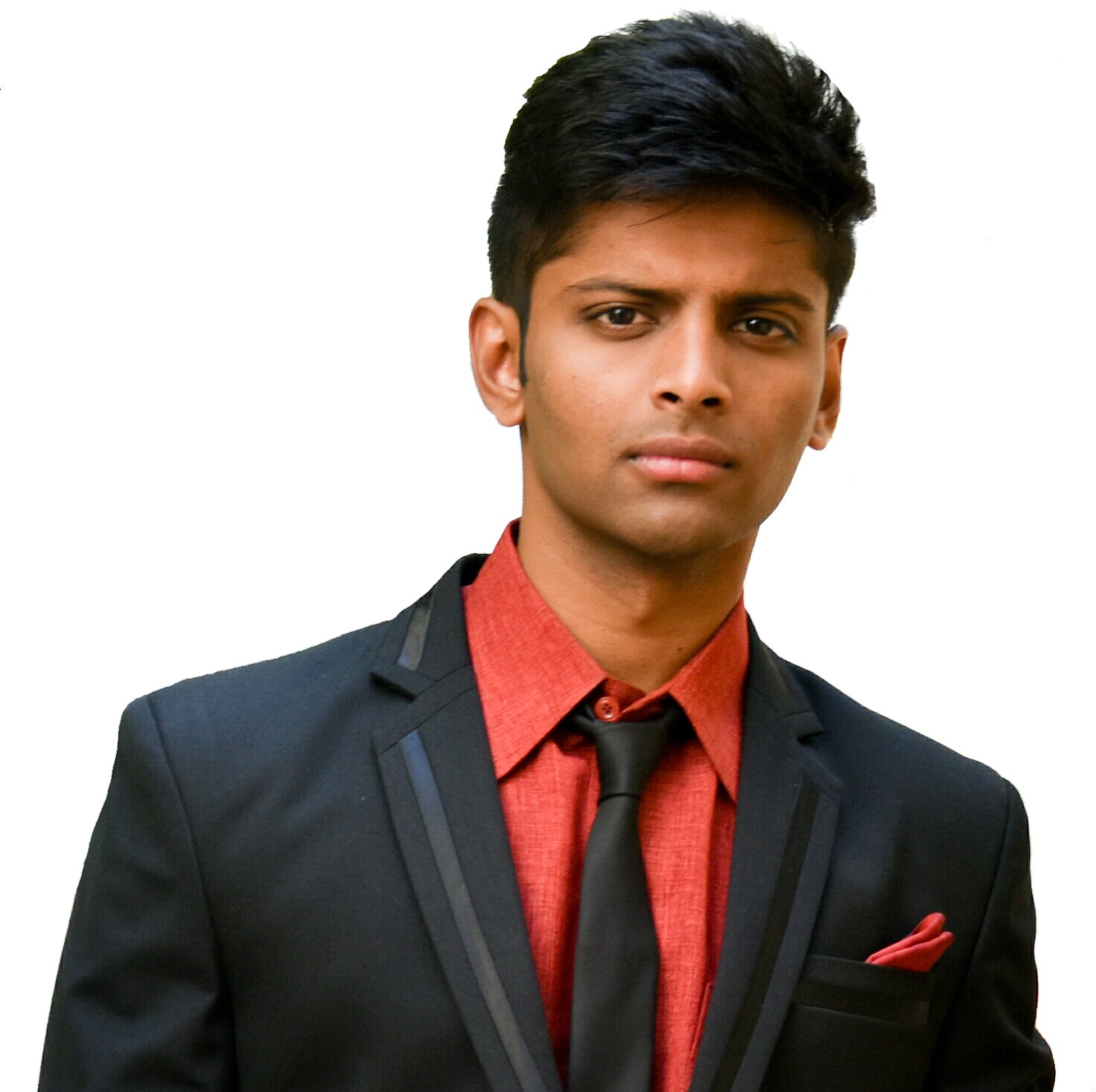}}]{Varun Ravi Kumar} received a B.E.\ degree in 2015 \& an M.Sc.\ degree in 2017 from TU Chemnitz, Germany. He is currently a Ph.D. student in Deep Learning for autonomous driving affiliated to TU Ilmenau and is currently working at Valeo. His research is mainly focused on the design of self-supervised perception algorithms using neural networks for self-driving cars. His diverse expertise lies in geometric, semantic tasks, 2D \& 3D object detection, point-cloud processing \& multi-task modeling. 
\end{IEEEbiography}
\vskip -2\baselineskip plus -1fil

\begin{IEEEbiography}[{\includegraphics[width=1in,height=1.25in,clip,keepaspectratio]{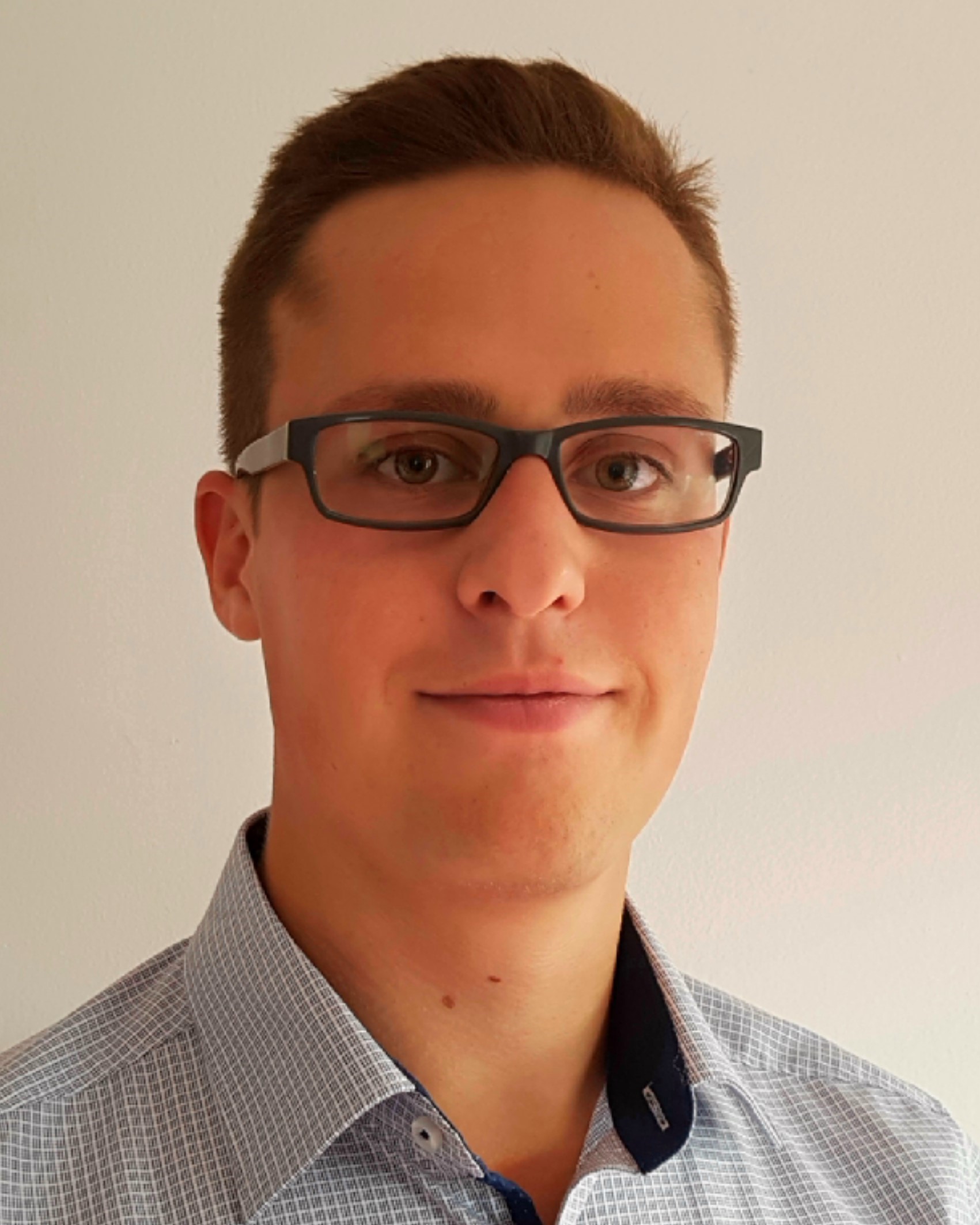}}]{Marvin Klingner}
received a B.Sc.\ degree in 2016 \& an M.Sc.\ degree in 2018 in physics from Georg-August-Universität Göttingen, Germany. He is currently a Ph.D.\ student in the Faculty of Electrical Engineering, Information Technology, Physics at Technische Universität Braunschweig, Germany.
His research interests in computer vision for autonomous driving include self-supervised 3D geometry perception with neural networks, multi-task learning approaches, the robustness of neural networks to domain shifts \& adversarial attacks \& performance monitoring neural networks during online deployment. He is the recipient of the Dr. Berliner - Dr. Ungewitter award of the Faculty of Physics at Georg-August-Universität Göttingen, Germany, in 2018, was given the CVPR SAIAD 2020 Workshop Best Paper Award.
\end{IEEEbiography}
\vskip -2\baselineskip plus -1fil

\begin{IEEEbiography}
[{\includegraphics[width=1in,height=1.25in,clip,keepaspectratio]{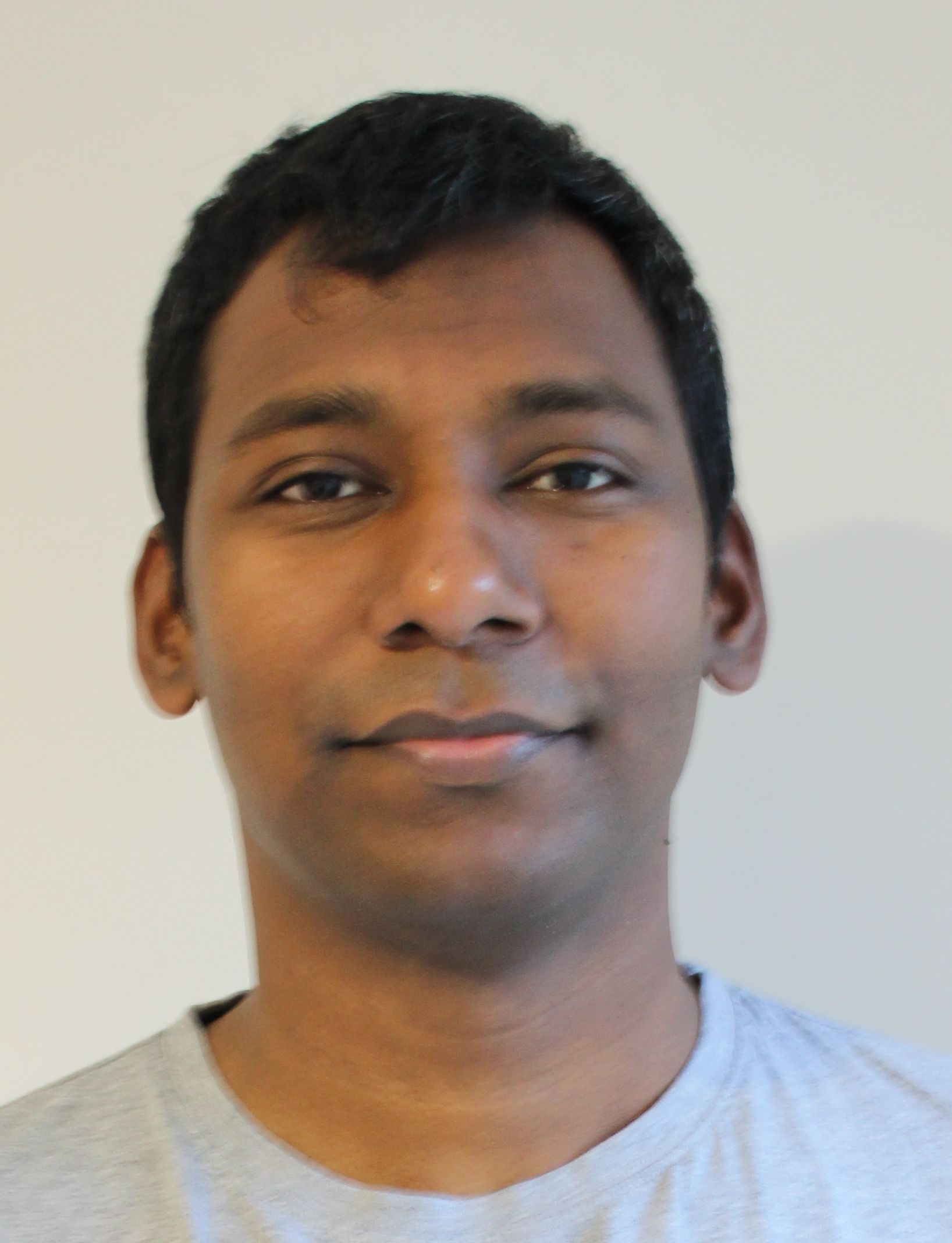}}]{Senthil Yogamani} is an Artificial Intelligence architect \& holds a director-level technical leader position at Valeo Ireland. He leads the research \& design of AI algorithms for various modules of autonomous driving systems. He has over 14 years of experience in computer vision \& machine learning, including 12 years of experience in industrial, automotive systems. He is an author of 100+ publications with 1600+ citations \& 100+ inventions with 70 filed patent families. He serves on the editorial board of various leading IEEE automotive conferences. 
\end{IEEEbiography}
\vskip -2\baselineskip plus -1fil

\begin{IEEEbiography}
[{\includegraphics[width=1in,height=1.25in,clip,keepaspectratio]{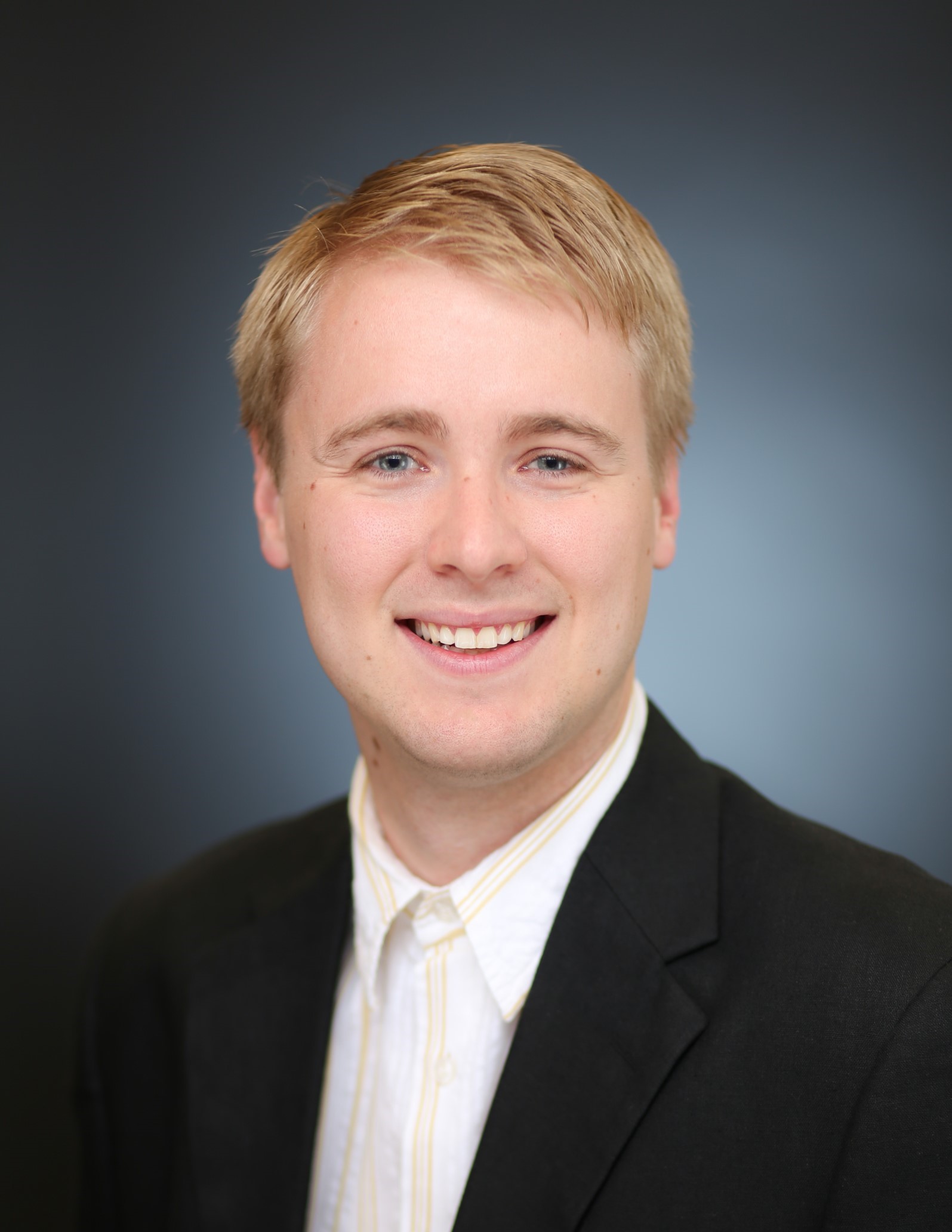}}]{Markus Bach} received the Dipl.-Phys. degree in 2013 from TU Dresden, Germany. He has over 5 years of research experience in the field of theoretical particle physics. In 2019 he joined Valeo as a software engineer in Driving Assistance Research \& is working on diverse topics: SLAM \& localization based on various inputs (camera, Lidar, Radar), computer vision, intrinsic \& extrinsic sensor calibration, assurance of AI-based perception functions for automated driving.
\end{IEEEbiography}
\vskip -2\baselineskip plus -1fil

\begin{IEEEbiography}
[{\includegraphics[width=1in,height=1.25in,clip,keepaspectratio]{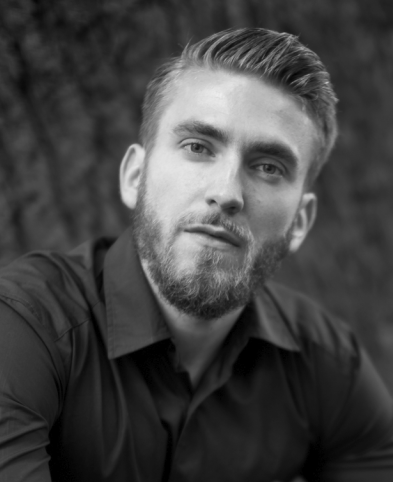}}]{Stefan Milz} received his Ph.D. degree in Physics from the Technical University of Munich. He has a strong history in professional software development \& automotive. He is Managing Director of Spleenlab.ai, a self-founded machine learning company focusing on safety-critical computer vision applications (UAV, Automated Driving, Air-Taxis) deploying SLAM, sensor-fusion, perception functions into the real world. Besides, he is also a research fellow at the TU-Ilmenau. Stefan Milz is the author \& co-author of more than 60 patents \& publications.
\end{IEEEbiography}
\vskip -2\baselineskip plus -1fil

\begin{IEEEbiography}[{\includegraphics[width=1in,height=1.25in,clip,keepaspectratio]{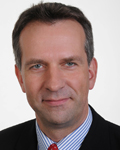}}]{Tim Fingscheidt}
received the Dipl.-Ing.\ degree in electrical engineering in 1993 \& the Ph.D.\ degree in 1998 from RWTH Aachen University, Germany.
He joined AT{\&}T Labs, Florham Park, NJ, USA, in 1998, \& Siemens AG (Mobile Devices), Munich, Germany, in 1999. With Siemens Corporate Technology, Munich, Germany, he led the speech technology development activities (2005–2006). Since 2006, he has been a Full Professor with the Institute for Communications Technology, Technische Universität Braunschweig, Germany. 
\end{IEEEbiography}
\vskip -2\baselineskip plus -1fil

\begin{IEEEbiography}
[{\includegraphics[width=1in,height=1.25in,clip,keepaspectratio]{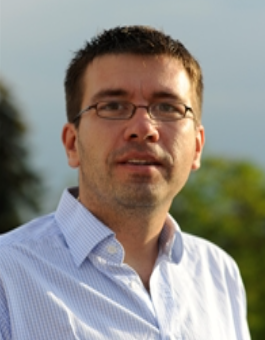}}]{Patrick Mäder} is a Professor at the Technische Universität Ilmenau, Germany \& heading the endowed chair on Software Engineering for Critical Systems. His research interests include software engineering focusing on requirements traceability, requirements engineering, object-oriented analysis \& design, \& development of safety-critical systems. He received a Diploma degree in industrial engineering \& a Ph.D. degree (Distinction) in computer science from the Technische Universität Ilmenau in 2003 \& 2009, respectively. 
\end{IEEEbiography}